%% file: conll2018.tex
\documentclass[11pt,a4paper]{article}
\usepackage[hyperref]{conll2018}
\usepackage{times}
\usepackage{latexsym}
\usepackage[T1]{fontenc}
\usepackage{url}

\usepackage{microtype}
\usepackage{graphicx}
\usepackage{subcaption}
\usepackage{booktabs} 
\usepackage{braket}
\usepackage{amsfonts,amsmath}
\usepackage{color}
\usepackage{mathtools}
\usepackage[toc,page]{appendix}
\usepackage{amsthm}
\usepackage{wrapfig}
\usepackage{float}
\usepackage{afterpage}

\usepackage{soul}

\setul{0.5ex}{0.3ex}
\setulcolor{red}

\newcommand*\rot{\rotatebox{90}}

\usepackage{pifont}
\newcommand{\cmark}{\ding{51}}%
\newcommand{\xmark}{\ding{55}}%
\usepackage{changepage}
\usepackage{float}
\usepackage[inline]{enumitem}
\newcommand{\redc}[1] {\textcolor{red}{#1}}
\newcommand{\bluec}[1] {\textcolor{blue}{#1}}
\newcommand{\greenc}[1] {\textcolor{green}{#1}}
\newcommand{\orangec}[1] {\textcolor{orange}{#1}}
\usepackage{longtable}

\renewcommand{\epsilon}{\ensuremath\varepsilon}

\renewcommand{\phi}{\ensuremath{\varphi}}

\aclfinalcopy 

\title{End-to-End Neural Entity Linking}

\author{Nikolaos Kolitsas \thanks{\ \ Equal contribution.} \\
  ETH Z\"urich \\
  {\tt nikos\_kolitsas@hotmail.com} \\
  \And
  Octavian-Eugen Ganea \footnotemark[1] \\
  ETH Z\"urich \\
  {\tt octavian.ganea@inf.ethz.ch} \\ 
  \AND
  Thomas Hofmann \\
  ETH Z\"urich \\
  {\tt thomas.hofmann@inf.ethz.ch} \\
}

\date{}

\begin{document}
\maketitle

\newcommand{\mat}[1]{{\mathbf #1}}
\newcommand{\x}{{\mathbf x}}
\newcommand{\I}{{\mathbf I}}
\newcommand{\hyperspace}{{\mathcal X}}
\newcommand{\D}{{\mathbb D}}
\newcommand{\HH}{{\mathbb H}}
\newcommand{\E}{{\mathbb E}}
\newcommand{\M}{{\mathbf M}}
\newcommand{\R}{{\mathbb R}}
\newcommand{\C}{{\mathbb C}}
\newcommand{\z}{{\mathbf z}}
\renewcommand{\Re}{{\mathbb R}}
\newcommand*{\QEDB}{\hfill\ensuremath{\square}}%
\newcommand{\hi}[1]{\textcolor{red}{#1}}

\begin{abstract}
Entity Linking (EL) is an essential task for semantic text understanding and information extraction. Popular methods separately address the Mention Detection (MD) and Entity Disambiguation (ED) stages of EL, without leveraging their mutual dependency. We here propose the first neural end-to-end EL system that jointly discovers and links entities in a text document. The main idea is to consider all possible spans as potential mentions and learn contextual similarity scores over their entity candidates that are useful for both MD and ED decisions. Key components are context-aware mention embeddings, entity embeddings and a probabilistic mention - entity map, without demanding other engineered features. Empirically, we show that our end-to-end method significantly outperforms popular systems on the Gerbil platform when enough training data is available. Conversely, if testing datasets follow different annotation conventions compared to the training set (e.g. queries/ tweets vs news documents), our ED model coupled with a traditional NER system offers the best or second best EL accuracy.
\end{abstract}

%
\input{introduction}

\input{related}

\input{model}

\input{experiments}

\input{conclusion}

\bibliography{mybibliography}
\bibliographystyle{acl_natbib_nourl}


\appendix

\input{appendix}
\end{document}

%% file: introduction.tex
\section{Introduction and Motivation}
Towards the goal of automatic text understanding, machine learning models are expected to accurately extract potentially ambiguous mentions of entities from a textual document and link them to a knowledge base (KB), e.g. Wikipedia or Freebase. Known as entity linking, this problem is an essential building block for various Natural Language Processing tasks, e.g. automatic KB construction, question-answering, text summarization, or relation extraction. 

\begin{table}[t]
\centering
\begin{tabular}{l} 
 \hline
\begin{tabular}{@{}c@{}}{\color{red}1) MD may split a larger span into two mentions of}\\{\color{red}less informative entities:} \end{tabular} \\ 
{\color{blue}\ul{B. Obama}'s \ul{wife}} gave a speech [...] \\
{\color{blue}\ul{Federer}'s \ul{coach}} [...]\\
\hline
\begin{tabular}{@{}c@{}}{\color{red}2) MD may split a larger span into two mentions of}\\{\color{red}incorrect entities:} \end{tabular} \\ 
{\color{blue}\ul{Obama} \ul{Castle}} was built in 1601 in Japan. \\
{\color{blue}The \ul{Kennel} \ul{Club}} is UK's official {\color{blue}kennel club}.\\ 
A {\color{blue}\ul{bird} \ul{dog}} is a type of {\color{blue}gun dog} or hunting dog. \\ 
{\color{blue}\ul{Romeo} and \ul{Juliet}} by Shakespeare [...] \\
{\color{blue}Natural \ul{killer} \ul{cells}} are a type of lymphocyte \\
{\color{blue}\ul{Mary} and \ul{Max}}, the 2009 movie [...]\\
\hline
\begin{tabular}{@{}c@{}}{\color{red}3) MD may choose a shorter span,}\\{\color{red}referring to an incorrect entity:} \end{tabular} \\ 
{\color{blue}The \ul{Apple}} is played again in cinemas. \\
{\color{blue}The \ul{New York} Times} is a popular newspaper. \\ 
\hline
\begin{tabular}{@{}c@{}}{\color{red}4) MD may choose a longer span,}\\{\color{red}referring to an incorrect entity:} \end{tabular} \\  
Babies \ul{{\color{blue}Romeo} and {\color{blue}Juliet}} were born hours apart.\\ 
\hline
\end{tabular}
\caption{Examples where MD may benefit from ED and viceversa. Each wrong MD decision (\ul{underlined}) can be avoided by proper context understanding. The correct spans are shown in {\color{blue}blue}.}
\label{tab:examples}
\vspace{-.2cm}
\end{table}

An EL system typically performs two tasks: i) Mention Detection (MD) or Named Entity Recognition (NER) when restricted to named entities -- extracts entity references in a raw textual input, and ii) Entity Disambiguation (ED) -- links these spans to their corresponding entities in a KB. Until recently, the common approach of popular systems~\citep{ceccarelli2013dexter,van2013learning,piccinno2014tagme,daiber2013improving,hoffart2011robust,steinmetz2013semantic} was to solve these two sub-problems independently. However, the important dependency between the two steps is ignored and errors caused by MD/NER will propagate to ED without possibility of recovery~\citep{sil2013re,luo2015joint}. We here advocate for models that address the end-to-end EL task, informally arguing that humans understand and generate text in a similar joint manner, discussing about entities which are gradually introduced, referenced under multiple names and evolving during time~\citep{ji2017dynamic}. Further, we emphasize the importance of the mutual dependency between MD and ED. First, numerous and more informative linkable spans found by MD obviously offer more contextual cues for ED. Second, finding the true entities appearing in a specific context encourages better mention boundaries, especially for multi-word mentions. For example, in the first sentence of Table~\ref{tab:examples}, understanding the presence of the entity \texttt{Michelle Obama} helps detecting its true mention \textit{"B. Obama's wife"}, as opposed to separately linking \textit{B. Obama} and \textit{wife} to less informative concepts.

We propose a simple, yet competitive, model for end-to-end EL. Getting inspiration from the recent works of \citep{lee2017end} and \citep{ganea2017deep}, our model first generates all possible spans (mentions) that have at least one possible entity candidate. Then, each  mention - candidate pair receives a context-aware compatibility score based on word and entity embeddings coupled with a neural attention and a global voting mechanisms. During training, we enforce the scores of gold entity - mention pairs to be higher than all possible scores of incorrect candidates or invalid mentions, thus jointly taking the ED and MD decisions. \\

Our contributions are:
\begin{itemize}
\item We address the end-to-end EL task using a simple model that conditions the "linkable" quality of a mention to the strongest context support of its best entity candidate. We do not require expensive manually annotated negative examples of non-linkable mentions. Moreover, we are able to train competitive models using little and only partially annotated documents (with named entities only such as the CoNLL-AIDA dataset). 
\item We are among the first to show that, with one single exception, engineered features can be fully replaced by neural embeddings automatically learned for the joint MD \& ED task. 
\item On the Gerbil\footnote{\url{http://gerbil.aksw.org/gerbil/}} benchmarking platform, we empirically show significant gains for the end-to-end EL task when test and training data come from the same domain. Morever, when testing datasets follow different annotation schemes or exhibit different statistics, our method is still effective in achieving state-of-the-art or close performance, but needs to be coupled with a popular NER system.
\end{itemize}

%% file: related.tex
\section{Related Work}

With few exceptions, MD/NER and ED are treated separately in the vast EL literature.

Traditional NER models usually view the problem as a word sequence labeling that is modeled using conditional random fields on top of engineered features~\citep{finkel2005incorporating} or, more recently, using bi-LSTMs architectures~\citep{lample2016neural,chiu2016named,liu2017empower} capable of learning complex lexical and syntactic features.

In the context of ED, recent neural methods~\citep{he2013learning,sun2015modeling,yamada2016joint,ganea2017deep,le2018improving,yang2018collective,radhakrishnan2018elden} have established state-of-the-art results, outperforming engineered features based models. Context aware word, span and entity embeddings, together with neural similarity functions, are essential in these frameworks.


End-to-end EL is the realistic task and ultimate goal, but challenges in joint NER/MD and ED modeling arise from their different nature. Few previous methods tackle the joint task, where errors in one stage can be recovered by the next stage. One of the first attempts,~\citep{sil2013re} use a popular NER model to over-generate mentions and let the linking step to take the final decisions. However, their method is limited by the dependence on a good mention spotter and by the usage of hand-engineered features. It is also unclear how linking can improve their MD phase. Later,~\citep{luo2015joint} presented one of the most competitive joint MD and ED models leveraging semi-Conditional Random Fields (semi-CRF). However, there are several weaknesses in this work. First, the mutual task dependency is weak, being captured only by type-category correlation features. The other engineered features used in their model are either NER or ED specific. Second, while their probabilistic graphical model allows for tractable learning and inference, it suffers from high computational complexity caused by the usage of the cartesian product of all possible document span segmentations, NER categories and entity assignments. Another approach is J-NERD~\citep{nguyen2016j} that addresses the end-to-end task using only engineered features and a probabilistic graphical model on top of sentence parse trees.


%% file: model.tex
\begin{figure*}[!htb]
\centering
\includegraphics[scale=0.25]{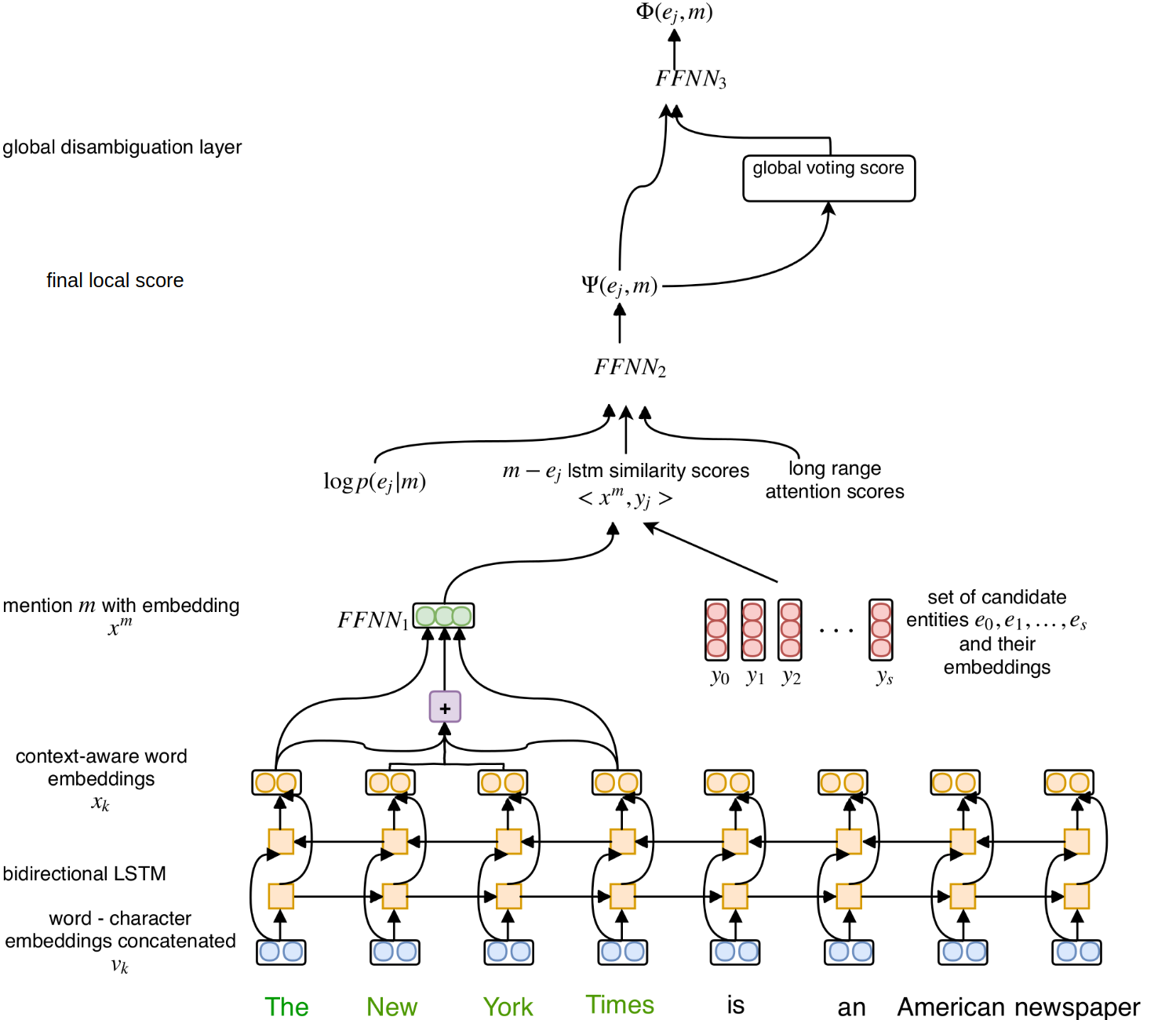}
\caption{Our global model architecture shown for the mention \textit{The New York Times}. The final score is used for both the mention linking and entity disambiguation decisions.}
\label{fig:model}
\end{figure*}

\section{Neural Joint Mention Detection and Entity Disambiguation}

We formally introduce the tasks of interest. For EL, the input is a text document (or a query or tweet) given as a sequence $D = \{ w_1, \ldots , w_n \} $ of words from a dictionary, $w_k \in \mathcal{W}$. The output of an EL model is a list of mention - entity pairs $\{ (m_i, e_i) \}_{i \in \overline{1,T}}$, where each mention is a word subsequence of the input document, $m = w_{q}, \ldots, w_{r}$, and each entity is an entry in a knowledge base KB (e.g. Wikipedia), $e \in \mathcal{E}$. For the ED task, the list of entity mentions $\{m_i\}_{i = \overline{1,T}}$ that need to be disambiguated is additionally provided as input. The expected output is a list of corresponding annotations $\{e_i\}_{i = \overline{1,T}} \in \mathcal{E}^T$. 

Note that, in this work, we only link mentions that have a valid gold KB entity, setting referred in~\citep{roder2017gerbil} as \textit{InKB} evaluation. Thus, we treat mentions referring to entities outside of the KB as "non-linkable". This is in line with few previous models, e.g.~\citep{luo2015joint,ganea2017deep,yamada2016joint}. We leave the interesting setting of discovering out-of-KB entities as future work.

We now describe the components of our neural end-to-end EL model, depicted in Figure~\ref{fig:model}. We aim for simplicity, but competitive accuracy. 

\paragraph{Word and Char Embeddings.}
We use pre-trained Word2Vec vectors~\citep{mikolov2013distributed}. In addition, we train character embeddings that capture important word lexical information. Following~\citep{lample2016neural}, for each word independently, we use bidirectional LSTMs~\citep{hochreiter1997long} on top of learnable char embeddings. These character LSTMs do not extend beyond single word boundaries, but they share the same parameters. Formally, let $\{z_1, \ldots , z_L \}$ be the character vectors of word $w$. We use the forward and backward LSTMs formulations defined recursively as in~\citep{lample2016neural}:
\begin{align}
&h^f_{t} = FWD-LSTM(h^f_{t-1}, z_t) \\ \nonumber
&h^b_{t} = BKWD-LSTM(h^b_{t+1}, z_t)
\end{align} 
Then, we form the character embedding of $w$ is $[h^f_{L}; h^b_{1}]$ from the hidden state of the forward LSTM corresponding to the last character concatenated with the hidden state of the backward LSTM corresponding to the first character. This is then concatenated with the pre-trained word embedding, forming the context-independent word-character embedding of $w$. We denote the sequence of these vectors as $\{v_k\}_{k \in \overline{1,n}}$ and depict it as the first neural layer in Figure~\ref{fig:model}.

\paragraph{Mention Representation.}
We find it crucial to make word embeddings aware of their local context, thus being informative for both mention boundary detection and entity disambiguation (leveraging contextual cues, e.g. "newspaper"). We thus encode context information into words using a bi-LSTM layer on top of the word-character embeddings $\{v_k\}_{k \in \overline{1,n}}$. The hidden states of forward and backward LSTMs corresponding to each word are then concatenated into \textit{context-aware word embeddings}, whose sequence is denoted as $\{x_k\}_{k \in \overline{1,n}}$. 

Next, for each possible mention, we produce a fixed size representation inspired by~\citep{lee2017end}. Given a mention $m = w_{q}, \ldots, w_{r}$, we first concatenate the embeddings of the first, last and the "soft head" words of the mention: 
\begin{align}
g^{m} = [x_q; x_r; \hat{x}^m]
\end{align}
The soft head embedding $\hat{x}^m$ is built using an attention mechanism on top of the mention's word embeddings, similar with~\citep{lee2017end}:
\begin{align}
& \alpha_k = \langle \boldsymbol{w}_{\alpha}, x_k \rangle \nonumber \\ 
& a^m_{k} = \frac{\exp(\alpha_k)}{\sum_{t=q}^r \exp(\alpha_t) } \\
& \hat{x}^m = \sum_{k=q}^r a^m_{k} \cdot v_k  \nonumber
\end{align}
However, we found the soft head embedding to only marginally improve results, probably due to the fact that most mentions are at most 2 words long. To learn non-linear interactions between the component word vectors, we project $g^m$ to a final mention representation with the same size as entity embeddings (see below) using a shallow feed-forward neural network FFNN (a simple projection layer):
\begin{align}
x^m = \text{FFNN}_1(g^{m})
\end{align}

\paragraph{Entity Embeddings.}
We use fixed continuous entity representations, namely the pre-trained entity embeddings of~\citep{ganea2017deep}, due to their simplicity and compatibility with the pre-trained word vectors of~\citep{mikolov2013distributed}. Briefly, these vectors are computed for each entity in isolation using the following exponential model that approximates the empirical conditional word-entity distribution $\hat{p}(w|e)$ obtained from co-occurrence counts. 
\begin{align}
\frac{\exp(\langle x_w, y_e \rangle)}{\sum_{w' \in \mathcal{W}} \exp(\langle x_{w'}, y_e \rangle)} \approx \hat{p}(w|e) 
\end{align}
Here, $x_w$ are fixed pre-trained word vectors and $y_e$ is the entity embedding to be trained. In practice,~\citep{ganea2017deep} re-write this as a max-margin objective loss.

\paragraph{Candidate Selection.}
For each span $m$ we select up to $s$ entity candidates that might be referred by this mention. These are top entities based on an empirical probabilistic entity - map $p(e|m)$ built by~\citep{ganea2017deep} from Wikipedia hyperlinks, Crosswikis~\citep{spitkovsky2012cross} and YAGO~\citep{hoffart2011robust} dictionaries. We denote by $C(m)$ this candidate set and use it both at training and test time.

\paragraph{Final Local Score.}
For each span $m$ that can possibly refer to an entity (i.e. $|C(m)| \geq 1$) and for each of its entity candidates $e_j \in C(m)$, we compute a similarity score using embedding dot-product that supposedly should capture useful information for both MD and ED decisions. We then combine it with the log-prior probability using a shallow FFNN, giving the context-aware entity - mention score:
\begin{align}
\Psi(e_j,m) = \text{FFNN}_2([\log p(e_j|m); \langle x^m, y_j\rangle])
\label{eq:psi}
\end{align}

\paragraph{Long Range Context Attention.}
In some cases, our model might be improved by explicitly capturing long context dependencies. To test this, we experimented with the attention model of~\citep{ganea2017deep}. This gives one context embedding per mention based on informative context words that are related to at least one of the candidate entities. We use this additional context embedding for computing dot-product similarity with any of the candidate entity embeddings. This value is fed as additional input of FFNN$_2$ in Eq.~\ref{eq:psi}. We refer to this model as \textit{long range context attention}.

\paragraph{Training.}
We assume a corpus with documents and gold entity - mention pairs $\mathcal{G} = \{(m_i,e_i^*)\}_{i = \overline{1,K}}$ is available. At training time, for each input document we collect the set $M$ of all (potentially overlapping) token spans $m$ for which $|C(m)| \geq 1$. We then train the parameters of our model using the following minimization procedure:
\begin{align}
\theta^* = \arg\min_{\theta} \sum_{m \in M} \sum_{e \in C(m)} V(\Psi_{\theta}(e,m))
\label{eq:loss}
\end{align}
where the violation term $V$ enforces the scores of gold pairs to be linearly separable from scores of negative pairs, i.e.
\begin{align}
& V(\Psi(e,m))= \\ \nonumber
& \quad \begin{cases}
    \max(0, \gamma - \Psi(e,m)) ,& \text{if} (e,m) \in \mathcal{G} \\
    \max(0, \Psi(e,m)) ,              & \text{otherwise}
\end{cases}
\end{align}
Note that, in the absence of annotated negative examples of "non-linkable" mentions, we assume that all spans in $M$ and their candidates that do not appear in $\mathcal{G}$ should not be linked. The model will be enforced to only output negative scores for all entity candidates of such mentions. 

We call \textit{all spans training} the above setting. Our method can also be used to perform ED only, in which case we train only on gold mentions, i.e. $M= \{m | m \in \mathcal{G}\}$. This is referred as \textit{gold spans training}. 

\paragraph{Inference.} At test time, our method can be applied for both \textit{EL} and \textit{ED only} as follows. First, for each document in our validation or test sets, we select all possibly linkable token spans, i.e. $M = \{m |\ |C(m)| \geq 1\}$ for EL, or the input set of mentions $M= \{m | m \in \mathcal{G}\}$ for ED, respectively. Second, the best linking threshold $\delta$ is found on the validation set such that the micro F1 metric is  maximized when only linking mention - entity pairs with $\Psi$ score greater than $\delta$. At test time, only entity - mention pairs with a score higher than $\delta$ are kept and sorted according to their $\Psi$ scores; the final annotations are greedily produced based on this set such that only spans not overlapping with previously selected spans (of higher scores) are chosen.

\paragraph{Global Disambiguation} \label{paragraph_global_disambiguation}
Our current model is "local", i.e. performs disambiguation of each candidate span independently. To enhance it, we add an extra layer to our neural network that will promote coherence among linked and disambiguated entities inside the same document, i.e. the \textit{global disambiguation layer}. Specifically, we compute a "global" mention-entity score based on which we produce the final annotations. We first define the set of mention-entity pairs that are allowed to participate in the global disambiguation voting, namely those that already have a high local score:
\begin{align*} 
V_{G} = \{(m,e) | m \in M, e \in C(m), \Psi(e, m) \geq \gamma' \}
\end{align*}
Since we initially consider all possible spans and for each span up to \textit{s} candidate entities, this filtering step is important  to avoid both undesired noise and exponential complexity for the EL task for which $M$ is typically much bigger than for ED. The final "global" score $G(e_j, m)$ for entity candidate $e_j$ of mention $m$ is given by the cosine similarity between the entity embedding and the normalized average of all other voting entities' embeddings (of the other mentions $m'$).
\begin{align}\label{global_voters}
& V_{G}^{m} = \{e | (m', e) \in V_{G} \land m' \neq m \}  \nonumber \\
& y_G^{m} = \sum\limits_{e \in V_{G}^{m} } y_e  \nonumber\\
& G(e_j, m) = \cos( y_{e_j}, y_G^{m}) \nonumber
\end{align} 
This is combined with the local score, yielding
$$\Phi(e_j,m) = \text{FFNN}_3([\Psi(e_j, m); G(e_j, m)]) $$
The final loss function is now slightly modified. Specifically, we enforce the linear separability in two places: in $\Psi(e,m)$ (exactly as before), but also in $\Phi(e,m)$, as follows
\begin{align} 
\theta^* & = \arg\min_{\theta} \\
							& \sum_{d \in D}  \sum_{m \in M} \sum_{e \in C(m)}  V(\Psi_{\theta}(e,m)) + V(\Phi_{\theta}(e,m)) \nonumber 
\label{eq:global_loss}
\end{align}

The inference procedure remains unchanged in this case, with the exception that it will only use the $\Phi(e,m)$ global score.

\paragraph{Coreference Resolution Heuristic. } In a few cases we found important to be able to solve simple coreference resolution cases (e.g. "Alan" referring to "Alan Shearer"). These cases are difficult to handle by our candidate selection strategy. We thus adopt the simple heuristic descried in~\citep{ganea2017deep} and observed between 0.5\% and 1\% improvement on all datasets.

%

%% file: experiments.tex
\section{Experiments}

\begin{table*}[!ht]
\scriptsize
\centering
\begin{tabular}{@{} l|c| c| c| c| c| c| c| c| c| c| c| c| c| c| @{}}
\multicolumn{1}{p{1cm}}{\textbf{F1@MA  F1@MI}}  &\rot{      AIDA A  } &\rot{      AIDA B  } &\rot{      MSNBC } &\rot{      OKE-2015} &\rot{OKE-2016} &\rot{      N3-Reuters-128   } &\rot{      N3-RSS-500 } &\rot{      Derczynski } &\rot{      KORE50 }\\
\hline
\hline
FREME& \multicolumn{1}{p{0.6cm}|}{23.6 37.6}& \multicolumn{1}{p{0.6cm}|}{23.8 36.3}& \multicolumn{1}{p{0.6cm}|}{15.8 19.9}& \multicolumn{1}{p{0.6cm}|}{26.1 31.6}& \multicolumn{1}{p{0.6cm}|}{22.7 28.5}& \multicolumn{1}{p{0.6cm}|}{26.8 30.9}& \multicolumn{1}{p{0.6cm}|}{32.5 27.8}& \multicolumn{1}{p{0.6cm}|}{31.4 18.9}& \multicolumn{1}{p{0.6cm}|}{12.3 14.5}\\
\hline
FOX& \multicolumn{1}{p{0.6cm}|}{54.7 58.0}& \multicolumn{1}{p{0.6cm}|}{58.1 57.0}& \multicolumn{1}{p{0.6cm}|}{11.2 8.3}& \multicolumn{1}{p{0.6cm}|}{53.9 56.8}& \multicolumn{1}{p{0.6cm}|}{49.5 50.5}& \multicolumn{1}{p{0.6cm}|}{\bluec{52.4 53.3}}& \multicolumn{1}{p{0.6cm}|}{35.1 33.8}& \multicolumn{1}{p{0.6cm}|}{42.0 38.0}& \multicolumn{1}{p{0.6cm}|}{28.3 30.8}\\
\hline
Babelfy& \multicolumn{1}{p{0.6cm}|}{41.2 47.2}& \multicolumn{1}{p{0.6cm}|}{42.4 48.5}& \multicolumn{1}{p{0.6cm}|}{36.6 39.7}& \multicolumn{1}{p{0.6cm}|}{39.3 41.9}& \multicolumn{1}{p{0.6cm}|}{37.8 37.7}& \multicolumn{1}{p{0.6cm}|}{19.6 23.0}& \multicolumn{1}{p{0.6cm}|}{32.1 29.1}& \multicolumn{1}{p{0.6cm}|}{28.9 29.8}& \multicolumn{1}{p{0.6cm}|}{\redc{52.5 55.9}}\\
\hline
Entityclassifier.eu& \multicolumn{1}{p{0.6cm}|}{43.0 44.7}& \multicolumn{1}{p{0.6cm}|}{42.9 45.0}& \multicolumn{1}{p{0.6cm}|}{41.4 42.2}& \multicolumn{1}{p{0.6cm}|}{29.2 29.5}& \multicolumn{1}{p{0.6cm}|}{33.8 32.5}& \multicolumn{1}{p{0.6cm}|}{24.7 27.9}& \multicolumn{1}{p{0.6cm}|}{23.1 22.7}& \multicolumn{1}{p{0.6cm}|}{16.3 16.9}& \multicolumn{1}{p{0.6cm}|}{25.2 28.0}\\
\hline
Kea& \multicolumn{1}{p{0.6cm}|}{36.8 40.4}& \multicolumn{1}{p{0.6cm}|}{39.0 42.3}& \multicolumn{1}{p{0.6cm}|}{30.6 30.9}& \multicolumn{1}{p{0.6cm}|}{44.6 46.2}& \multicolumn{1}{p{0.6cm}|}{46.3 46.4}& \multicolumn{1}{p{0.6cm}|}{17.5 18.1}& \multicolumn{1}{p{0.6cm}|}{22.7 20.5}& \multicolumn{1}{p{0.6cm}|}{31.3 26.5}& \multicolumn{1}{p{0.6cm}|}{41.0 46.8}\\
\hline
DBpedia Spotlight& \multicolumn{1}{p{0.6cm}|}{49.9 55.2}& \multicolumn{1}{p{0.6cm}|}{52.0 57.8}& \multicolumn{1}{p{0.6cm}|}{42.4 40.6}& \multicolumn{1}{p{0.6cm}|}{42.0 44.4}& \multicolumn{1}{p{0.6cm}|}{41.4 43.1}& \multicolumn{1}{p{0.6cm}|}{21.5 24.8}& \multicolumn{1}{p{0.6cm}|}{26.7 27.2}& \multicolumn{1}{p{0.6cm}|}{33.7 32.2}& \multicolumn{1}{p{0.6cm}|}{29.4 34.9}\\
\hline
AIDA& \multicolumn{1}{p{0.6cm}|}{68.8 72.4}& \multicolumn{1}{p{0.6cm}|}{71.9 72.8}& \multicolumn{1}{p{0.6cm}|}{62.7 65.1}& \multicolumn{1}{p{0.6cm}|}{\bluec{58.7 63.1}}& \multicolumn{1}{p{0.6cm}|}{0.0 0.0}& \multicolumn{1}{p{0.6cm}|}{42.6 46.4}& \multicolumn{1}{p{0.6cm}|}{42.6 \redc{42.4}}& \multicolumn{1}{p{0.6cm}|}{40.6 32.6}& \multicolumn{1}{p{0.6cm}|}{\bluec{49.6 55.4}}\\
\hline
WAT& \multicolumn{1}{p{0.6cm}|}{69.2 72.8}& \multicolumn{1}{p{0.6cm}|}{70.8 73.0}& \multicolumn{1}{p{0.6cm}|}{62.6 64.5}& \multicolumn{1}{p{0.6cm}|}{53.2 56.4}& \multicolumn{1}{p{0.6cm}|}{\bluec{51.8 53.9}}& \multicolumn{1}{p{0.6cm}|}{45.0 49.2}& \multicolumn{1}{p{0.6cm}|}{45.3 42.3}& \multicolumn{1}{p{0.6cm}|}{44.4 38.0}& \multicolumn{1}{p{0.6cm}|}{37.3 49.6}\\
\hline
\begin{tabular}{@{}l@{}} \textbf{Best baseline} \end{tabular}  & \multicolumn{1}{p{0.6cm}|}{\textbf{69.2 72.8}}& \multicolumn{1}{p{0.6cm}|}{\textbf{71.9 73.0}}& \multicolumn{1}{p{0.6cm}|}{\textbf{62.7 65.1}}& \multicolumn{1}{p{0.6cm}|}{\textbf{58.7 63.1}}& \multicolumn{1}{p{0.6cm}|}{\textbf{51.8 53.9}}& \multicolumn{1}{p{0.6cm}|}{\textbf{52.4 53.3}}& \multicolumn{1}{p{0.6cm}|}{\textbf{45.3 42.4}}& \multicolumn{1}{p{0.6cm}|}{\textbf{44.4 38.0}}& \multicolumn{1}{p{0.6cm}|}{\textbf{52.5 55.9}}\\
\hline
base model& \multicolumn{1}{p{0.6cm}|}{\bluec{86.6 89.1}}& \multicolumn{1}{p{0.6cm}|}{81.1 80.5}& \multicolumn{1}{p{0.6cm}|}{64.5 65.7}& \multicolumn{1}{p{0.6cm}|}{54.3 58.2}& \multicolumn{1}{p{0.6cm}|}{43.6 46.0}& \multicolumn{1}{p{0.6cm}|}{47.7 49.0}& \multicolumn{1}{p{0.6cm}|}{44.2 38.8}& \multicolumn{1}{p{0.6cm}|}{43.5 38.1}& \multicolumn{1}{p{0.6cm}|}{34.9 42.0}\\
\hline
base model + att& \multicolumn{1}{p{0.6cm}|}{86.5 88.9}& \multicolumn{1}{p{0.6cm}|}{\bluec{81.9 82.3}}& \multicolumn{1}{p{0.6cm}|}{69.4 69.5}& \multicolumn{1}{p{0.6cm}|}{56.6 60.7}& \multicolumn{1}{p{0.6cm}|}{49.2 51.6}& \multicolumn{1}{p{0.6cm}|}{48.3 51.1}& \multicolumn{1}{p{0.6cm}|}{\redc{46.0} 40.5}& \multicolumn{1}{p{0.6cm}|}{\bluec{47.9} \redc{42.3}} & \multicolumn{1}{p{0.6cm}|}{36.0 42.2} \\
\hline
base model + att + global& \multicolumn{1}{p{0.6cm}|}{\redc{86.6 89.4}}& \multicolumn{1}{p{0.6cm}|}{\redc{82.6 82.4}}& \multicolumn{1}{p{0.6cm}|}{\redc{73.0 72.4}}& \multicolumn{1}{p{0.6cm}|}{56.6 61.9}& \multicolumn{1}{p{0.6cm}|}{47.8 52.7}& \multicolumn{1}{p{0.6cm}|}{45.4 50.3}& \multicolumn{1}{p{0.6cm}|}{43.8 38.2}& \multicolumn{1}{p{0.6cm}|}{43.2 34.1} & \multicolumn{1}{p{0.6cm}|}{26.2 35.2}\\
\hline
\begin{tabular}{@{}c@{}} ED base model + att + global using \\ Stanford NER mentions\end{tabular} & \multicolumn{1}{p{0.6cm}|}{75.7 80.3}& \multicolumn{1}{p{0.6cm}|}{73.3 74.6}& \multicolumn{1}{p{0.6cm}|}{ \bluec{71.1 71.0}}& \multicolumn{1}{p{0.6cm}|}{\redc{62.9} \redc{66.9}}& \multicolumn{1}{p{0.6cm}|}{\redc{57.1} \redc{58.4}}& \multicolumn{1}{p{0.6cm}|}{\redc{54.2 54.6}}& \multicolumn{1}{p{0.6cm}|}{\bluec{45.9 42.2}}& \multicolumn{1}{p{0.6cm}|}{\redc{48.8} \redc{42.3}}& \multicolumn{1}{p{0.6cm}|}{40.3 46.0}\\
\hline
\end{tabular}
\caption{EL strong matching results on the Gerbil platform. Micro and Macro F1 scores are shown. We highlight \redc{the best} and \bluec{second best} models, respectively. Training was done on AIDA-train set.}
\label{tab:el_strong}
\end{table*}

\paragraph{Datasets and Metrics.}
We used Wikipedia 2014 as our KB. We conducted experiments on the most important public EL datasets using the Gerbil platform~\citep{roder2017gerbil}. Datasets' statistics are provided in Tables~\ref{tab:aidastats} and~\ref{tab:datasets} (Appendix). This benchmarking framework offers reliable and trustable evaluation and comparison with state of the art EL/ED methods on most of the public datasets for this task. It also shows how well different systems generalize to datasets from very different domains and annotation schemes compared to their training sets. Moreover, it offers evaluation metrics for the end-to-end EL task, as opposed to some works that only evaluate NER and ED separately, e.g.~\citep{luo2015joint}.

As previously explained, we do not use the NIL mentions (without a valid KB entity) and only compare against other systems using the InKB metrics. 

For training we used the biggest publicly available EL dataset, AIDA/CoNLL~\citep{hoffart2011robust}, consisting of a training set of 18,448 linked mentions in 946 documents, a validation set of 4,791 mentions in 216 documents, and a test set of 4,485 mentions in 231 documents. 

We report micro and macro InKB F1 scores for both EL and ED. For EL, these metrics are computed both in the \textit{strong matching} and \textit{weak matching} settings. The former requires exactly predicting the gold mention boundaries and their entity annotations, whereas the latter gives a perfect score to spans that just overlap with the gold mentions and are linked to the correct gold entities.

\begin{table*}[!h]
\scriptsize
\centering
\begin{tabular}{@{} l|c| c| c| c|| c| c| c| c|@{}}
\textbf{\begin{tabular}{@{}c@{}}\multicolumn{1}{c}{solved\%}\\\multicolumn{1}{c}{matched gold entity\%}\\ \\position of ground\\truth in $p(e|m)$\end{tabular}} &\rot{\begin{tabular}{@{}c@{}}Number of\\mentions\end{tabular}}  &\rot{ED Global model} &\rot{ED Base + Att model} &\rot{ED Base model} &\rot{EL Global model} &\rot{EL Base + Att model} &\rot{EL Base model} &\rot{EL Global -log$p(e|m)$}\\
\hline
\hline
1& 3390 & \multicolumn{1}{p{0.6cm}|}{98.8 98.8}& \multicolumn{1}{p{0.6cm}|}{98.6 98.9}& \multicolumn{1}{p{0.6cm}||}{98.3 98.8}& \multicolumn{1}{p{0.6cm}|}{\redc{96.7 99.0}}& \multicolumn{1}{p{0.6cm}|}{96.6 98.6}& \multicolumn{1}{p{0.6cm}|}{96.2 99.0} & \multicolumn{1}{p{0.6cm}|}{93.3 96.7}\\
\hline
2& 655& \multicolumn{1}{p{0.6cm}|}{89.9 89.9}& \multicolumn{1}{p{0.6cm}|}{88.1 88.5}& \multicolumn{1}{p{0.6cm}||}{88.5 88.9}& \multicolumn{1}{p{0.6cm}|}{\redc{86.8 90.8}}& \multicolumn{1}{p{0.6cm}|}{86.8 90.8}& \multicolumn{1}{p{0.6cm}|}{85.0 88.4} & \multicolumn{1}{p{0.6cm}|}{86.9 89.8}\\
\hline
3& 108& \multicolumn{1}{p{0.6cm}|}{83.3 84.3}& \multicolumn{1}{p{0.6cm}|}{81.5 82.4}& \multicolumn{1}{p{0.6cm}||}{75.9 78.7}& \multicolumn{1}{p{0.6cm}|}{79.1 84.5}& \multicolumn{1}{p{0.6cm}|}{80.2 84.7}& \multicolumn{1}{p{0.6cm}|}{74.8 81.1} & \multicolumn{1}{p{0.6cm}|}{\redc{84.3 88.9}}\\
\hline
4-8& 262& \multicolumn{1}{p{0.6cm}|}{78.2 78.2}& \multicolumn{1}{p{0.6cm}|}{76.3 79.0}& \multicolumn{1}{p{0.6cm}||}{74.8 76.0}& \multicolumn{1}{p{0.6cm}|}{69.5 78.2}& \multicolumn{1}{p{0.6cm}|}{68.8 78.7}& \multicolumn{1}{p{0.6cm}|}{68.7 76.0} & \multicolumn{1}{p{0.6cm}|}{\redc{78.9 83.5}}\\
\hline
9+& 247& \multicolumn{1}{p{0.6cm}|}{59.9 63.6}& \multicolumn{1}{p{0.6cm}|}{53.4 60.7}& \multicolumn{1}{p{0.6cm}||}{53.0 58.7}& \multicolumn{1}{p{0.6cm}|}{47.8 58.2}& \multicolumn{1}{p{0.6cm}|}{46.2 59.4}& \multicolumn{1}{p{0.6cm}|}{50.4 54.8} & \multicolumn{1}{p{0.6cm}|}{\redc{62.7 67.5}}\\
\hline
\end{tabular}
\caption{AIDA A dataset: Gold mentions are split by the position they appear in the $p(e|m)$ dictionary. In each cell, the upper value is the percentage of the gold mentions that were annotated with the correct entity (recall), whereas the lower value is the percentage of gold mentions for which our system's highest scored entity is the ground truth entity, but that might not be annotated in the end because its score is below the threshold $\delta$.
}
\label{tab:gm_bucketing}
\end{table*}

\paragraph{Baselines.}
We compare with popular and state-of-the-art EL and ED public systems, e.g. WAT~\citep{piccinno2014tagme}, Dbpedia Spotlight~\citep{mendes2011dbpedia}, NERD-ML~\citep{van2013learning}, KEA~\citep{steinmetz2013semantic}, AIDA~\citep{hoffart2011robust}, FREME, FOX~\citep{speck2014ensemble}, PBOH~\citep{ganea2016probabilistic} or Babelfy~\citep{moro2014entity}. These are integrated within the Gerbil platform. However, we did not compare with~\citep{luo2015joint} and other models outside Gerbil that do not use end-to-end EL metrics.

\paragraph{Training Details and Model Hyper-Parameters.} We use the same model hyper-parameters for both EL and ED training. The only difference between the two settings is the span set $M$ : EL uses \textit{all spans training}, whereas ED uses \textit{gold spans training} settings as explained before.

Our pre-trained word and entity embeddings are 300 dimensional, while 50 dimensional trainable character vectors were used. The char LSTMs have also hidden dimension of 50. Thus, word-character embeddings are 400 dimensional. The contextual LSTMs have hidden size of 150, resulting in 300 dimensional context-aware word vectors. We apply dropout on the concatenated word-character embeddings, on the output of the bidirectional context LSTM and on the entity embeddings used in Eq.~\ref{eq:psi}. The three FFNNs in our model are simple projections without hidden layers (no improvements were obtained with deeper layers). For the \textit{long range context attention} we used a word window size of K = 200 and keep top R = 10 words after the hard attention layer (notations from~\citep{ganea2017deep}). We use at most $s=30$ entity candidate per mention both at train and test time. $\gamma$ is set to 0.2 without further investigations. $\gamma'$ is set to 0, but a value of 0.1 was giving similar results.

For the loss optimization we use Adam~\citep{kingma2014adam} with a learning rate of 0.001. We perform early stopping by evaluating the model on the AIDA validation set each 10 minutes and stopping after 6 consecutive evaluations with no significant improvement in the macro F1 score.

%

\begin{table*}[!h]
\centering
\begin{tabular}{p{15.5cm}} 
\hline
\textbf{1) Annotated document:} \\
SOCCER - \greenc{[}SHEARER \greenc{]} NAMED AS \redc{[1} \orangec{[1}ENGLAND\redc{]}  CAPTAIN\orangec{]}. \greenc{[} LONDON \greenc{]} 1996-08-30 The world 's costliest footballer \greenc{[} Alan Shearer \greenc{]} was named as the new \greenc{[}England \greenc{]} captain on Friday. The 26-year-old, who joined \greenc{[}Newcastle\greenc{]} for 15 million pounds sterling, takes over from \greenc{[}Tony Adams\greenc{]}, who led the side during the \greenc{[} European \greenc{]} championship in June, and former captain \greenc{[} David Platt \greenc{]} . \greenc{[2}Adams\greenc{]} and \greenc{[} Platt \greenc{]} are both injured and will miss \greenc{[}England\greenc{]}'s opening \redc{[3}\orangec{[}World Cup\redc{]}\orangec{]} qualifier against \greenc{[}Moldova\greenc{]} on Sunday . \greenc{[}Shearer\greenc{]} takes the captaincy on a trial basis , but new coach \greenc{[}Glenn Hoddle\greenc{]} said he saw no reason why the former \greenc{[}Blackburn\greenc{]} and \greenc{[}Southampton\greenc{]} skipper should not make the post his own . "I 'm sure \greenc{[4}Alan\greenc{]} is the man for the job , " \greenc{[}Hoddle\greenc{]} said . [...] I spoke to \greenc{[5}Alan\greenc{]} he was up for it [...]. \greenc{[}Shearer\greenc{]} 's \greenc{[}Euro 96\greenc{]} striking partner [...]. \\
\hline
\textbf{Analysis}:\\
\redc{[1} is considered a false negative and the ground truth is the entity England\_national\_football\_team. Our annotation was for the span "ENGLAND CAPTAIN" and wrongly linked to the England\_cricket\_team. \\
\redc{[3} The ground truth here is 1998\_FIFA\_World\_Cup whereas our model links it to FIFA\_World\_Cup.\\
\greenc{[2,4,5} are correctly solved due to our coreference resolution heuristic. \\
\hline

\hline
\textbf{2) Annotated document:} \\
\greenc{[} N. Korea \greenc{]} urges \greenc{[} S. Korea \greenc{]} to return war veteran . \greenc{[} SEOUL \greenc{]} 1996-08-31 \greenc{[} North Korea \greenc{]} demanded on Saturday that \greenc{[} South Korea \greenc{]} return a northern war veteran who has been in the \redc{[1} South \redc{]} since the 1950-53 war , \greenc{[} Seoul \greenc{]} 's unification ministry said . " ...I request the immediate repatriation of Kim In-so to \greenc{[} North Korea \greenc{]} where his family is waiting , " \orangec{[1} North Korean \orangec{]} Red Cross president Li Song-ho said in a telephone message to his southern couterpart , \orangec{[2} Kang Young-hoon \orangec{]} . Li said Kim had been critically ill with a cerebral haemorrhage . The message was distributed to the press by the \greenc{[} South Korean \greenc{]} unification ministry . Kim , an unrepentant communist , was captured during the \redc{[2} \orangec{[3} Korean \redc{]} War \orangec{]} and released after spending more than 30 years in a southern jail . He submitted a petition to the \greenc{[} International Red Cross \greenc{]} in 1993 asking for his repatriation . The domestic \greenc{[} Yonhap \greenc{]} news agency said the \greenc{[} South Korean \greenc{]} government would consider the northern demand only if the \redc{[3} North \redc{]} accepted \greenc{[} Seoul \greenc{]} 's requests , which include regular reunions of families split by the \redc{[4} \orangec{[4} Korean \redc{]} War \orangec{]} . Government officials were not available to comment . \greenc{[} South Korea \greenc{]} in 1993 unconditionally repatriated Li In-mo , a nothern partisan seized by the \redc{[5} South \redc{]} during the war and jailed for more than three decades.\\
\textbf{Analysis}:\\
\redc{[1, [3, and [5} cases illustrate the main source of errors. These are false negatives in which our model has the correct ground truth entity pair as the highest scored one for that mention, but since it is not confident enough (score $< \gamma$) it decides not to annotate that mention. In this specific document these errors could probably be avoided easily with a better coreference resolution mechanism.\\
\orangec{[3 and [4} cases illustrate that the gold standard can be problematic. Specifically, instead of annotating the whole span \textit{Korean War} and linking it to the war of 1950, the gold annotation only include \textit{Korean} and link it to the general entity of \textit{Korea\_(country)}.\\
\orangec{[2} is correctly annotated by our system but it is not included in the gold standard.\\
\hline

\end{tabular}
\caption{Error analysis on a sample document. Green corresponds to true positive (correctly discovered and annotated mention), red to false negative (ground truth mention or entity that was not annotated) and orange to false positive (incorrect mention or entity annotation ).}
\label{tab:error_analysis}
\end{table*}

\paragraph{Results and Discussion.}

The following models are used. 

i) \textbf{Base model}: only uses the mention local score and the log-prior. It does not use long range attention, nor global disambiguation. It does not use the head attention mechanism. 

ii) \textbf{Base model + att}: the Base Model plus Long Range Context Attention. 

ii) \textbf{Base model + att + global}: our Global Model (depicted in figure \ref{fig:model}) 

iv) \textbf{ED base model + att + global Stanford NER}: our ED Global model that runs on top of the detected mentions of the Stanford NER system~\citep{finkel2005incorporating}.

EL strong and weak matching results are presented in Tables~\ref{tab:el_strong} and ~\ref{tab:el_weak} (Appendix).

We first note that our system outperforms all baselines on the end-to-end EL task on both AIDA-A (dev) and AIDA-B (test) datasets, which are the biggest EL datasets publicly available. Moreover, we surpass all competitors on both EL and ED by a large margin, at least 9\%, showcasing the effectiveness of our method. We also outperform systems that optimize MD and ED separately, including our \textit{ED base model + att + global Stanford NER}. This demonstrates the merit of joint MD + ED optimization. 

In addition, one can observe that weak matching EL results are comparable with the strong matching results, showcasing that our method is very good at detecting mention boundaries.

At this point, our main goal was achieved: if enough training data is available with the same characteristics or annotation schemes as the test data, then our joint EL offers the best model. This is true not only when training on AIDA, but also for other types of datasets such as queries (Table~\ref{tab:gerdaq_results}) or tweets (Table~\ref{tab:micropost_results}). However, when testing data  has different statistics or follows different conventions than the training data, our method is shown to work best in conjunction with a state-of-the-art NER system as it can be seen from the results of our \mbox{\textbf{ED base model + att + global Stanford NER}} for different datasets in Table~\ref{tab:el_strong}. It is expected that such a NER system designed with a broader generalization scheme in mind would help in this case, which is confirmed by our results on different datasets.

While the main focus of this paper is the end-to-end EL and not the ED-only task, we do show ED results in Tables~\ref{tab:ed} and ~\ref{tab:ed_local_results}. We observe that our models are slightly behind recent top performing systems, but our unified EL - ED architecture has to deal with other challenges, e.g. being able to exchange global information between many more mentions at the EL stage, and is thus not suitable for expensive global ED strategies. We leave bridging this gap for ED as future work.

Additional results and insights are shown in the Appendix.

\paragraph{Ablation study}
Table~\ref{tab:gm_bucketing} shows an ablation study of our method. One can see that the $\log p(e|m)$ prior is very helpful for correctly linking unambiguous mentions, but is introducing noise when gold entities are not frequent. For this category of rare entities, removing this prior completely will result in a significant improvement, but this is not a practical choice since the gold entity is unknown at test time.

\paragraph{Error Analysis.}
We conducted a qualitative experiment shown in Table~\ref{tab:error_analysis}. We showcase correct annotations, as well as errors done by our system on the AIDA datasets. Inspecting the output annotations of our EL model, we discovered the remarkable property of not over-generating incorrect mentions, nor under-generating (missing) gold spans. We also observed that additional mentions generated by our model do correspond in the majority of time to actual KB entities, but are incorrectly forgotten from the gold annotations.

%% file: conclusion.tex
\section{Conclusion}
We presented the first neural end-to-end entity linking model and show the benefit of jointly optimizing entity recognition and linking. Leveraging key components, namely word, entity and mention embeddings, we prove that engineered features can be almost completely replaced by modern neural networks. Empirically, on the established Gerbil benchmarking platform, we exhibit state-of-the-art performance for EL on the biggest public dataset, AIDA/CoNLL, also showing good generalization ability on other datasets with very different characteristics when combining our model with the popular Stanford NER system.  

Our code is publicly available\footnote{\url{https://github.com/dalab/end2end_neural_el}}.

%% file: appendix.tex
\begin{table*}[!ht]
\scriptsize
\begin{center}
	\begin{tabular}{@{} c|c| c| c @{}}
	\textbf{\begin{tabular}{@{}l@{}}position of ground\\truth in $p(e|m)$\end{tabular}} & \textbf{AIDA Train} & \textbf{AIDA A (dev)} &\textbf{AIDA B (test)} \\
	\hline
	\hline
	1& \shortstack{13704\\75.6\%}& \shortstack{3390\\72.7\%}& \shortstack{3078\\ 69.8\%}\\
	\hline
	2& \shortstack{2339\\12.9\%}& \shortstack{655\\14\%}& \shortstack{651\\14.8\%}\\
	\hline
	3& \shortstack{565\\3.1\%}& \shortstack{108\\2.3\%}& \shortstack{184\\4.2\%}\\
	\hline
	4-8& \shortstack{843\\4.6\%}& \shortstack{262\\5.6\%}& \shortstack{308\\7\%}\\
	\hline
	9+& \shortstack{686\\3.8\%}& \shortstack{247\\5.3\%}& \shortstack{187\\4.2\%}\\
	\hline 
	\hline
	Total number of non-NIL mentions & 18137&	4662&	4408\\
	\hline
	\hline
	\begin{tabular}{@{}c@{}}Unambiguous mentions, i.e. \\for which ground truth is \\ the unique candidate\end{tabular}& \shortstack{3575\\19.7\%}& \shortstack{936\\20.1\%}& \shortstack{833\\18.9\%}
	\end{tabular}
\caption{Statistics on the number of mentions for various settings and for all AIDA datasets.}
\label{tab:aidastats}
\end{center}
\end{table*}

\begin{table*}[!ht]

  \begin{center}

    \begin{tabular}{c| c| c| c| c| c| c| c| c| r}    

      \rot{\textbf{Dataset}} & \rot{\textbf{Topic}} & \rot{\textbf{Type of annotations}} & \rot{\textbf{Exhaustive annotations}} & \rot{\textbf{Num Docs}} & \rot{\textbf{Avg. Words/Doc.}} & \rot{\textbf{Num InKB Anns}} & \rot{\textbf{\begin{tabular}{@{}c@{}} Num of gold entities\\ not in our KB\end{tabular}}} & \rot{\textbf{Recall 30 (\%)}} & \rot{\textbf{Recall 10 (\%)}}\\

      \hline    

      AIDA 		& news & NE 	& \cmark & 231 & 190 & 4483 & 5 & 98.3 & 95.7 \\
      \hline    

      AQUAINT 	& news & NE\&N  & \xmark & 50  & 221 & 727 & 16 & 93.5 & 92.6 \\
      \hline    

      MSNBC 	& news & NE 	& \cmark & 20  & 544 & 653 & 3  & 92.8 & 90.3 \\
      \hline    

      ACE2004	& news & NE 	& \xmark & 57  & 374 & 257 & 2 & 87.9 & 86.8 \\
      \hline    

	  \begin{tabular}{@{}c@{}}Derczynski \\\citep{derczynski2015analysis}\end{tabular} & tweets & NE 	& \cmark & 182 & 21  & 210 & 32 & 71.0 & 69.5 \\
      \hline    

      \begin{tabular}{@{}c@{}}Microposts2016 dev \\\citep{rizzo2014benchmarking}\end{tabular} & tweets & NE  & \cmark  & 100  & 14 & 251 & 50 & 59.8 & 59.8 \\
      \hline    

      N3-Reuters-128 & news & NE & \cmark & 128 & 124 & 631 & 7 & 68.9 & 64.3  \\
      \hline    

      N3-RSS-500 & news & NE & \xmark & 500 & 31 & 519 & 11 & 81.3 & 79.4 \\
      \hline    

      \begin{tabular}{@{}c@{}}DBpedia Spotlight\\\citep{mendes2011dbpedia}\end{tabular} & news & NE\&N & \cmark & 58 & 29 & 330 & 7 & 90.6 & 90.3 \\
      \hline    

      \begin{tabular}{@{}c@{}}KORE 50\\\citep{hoffart2012kore}\end{tabular} 	& mixed & NE & \cmark & 50 & 13 & 144 & 1 & 88.2 & 75.7  \\
      \hline    

      \begin{tabular}{@{}c@{}}ERD2014 \\\citep{carmel2014erd}\end{tabular} & queries & NE & \cmark & 91 & 3.5 & 59 & 1 & 72.9 & 72.9  \\
      \hline    

      \begin{tabular}{@{}c@{}}Gerdaq \\\citep{cornolti2016piggyback}\end{tabular} & queries & NE\&N & \cmark & 250 & 3.6 & 408 & 3 & 70.8 & 68.6 \\
      \hline    

      \begin{tabular}{@{}c@{}}OKE \\\citep{nuzzolese2015open}\end{tabular} & wikipedia & NE\&roles & \cmark & 55 & 26.6 & 288 & 3 & 81.9 & 81.3  \\

    \end{tabular}

  \caption{Test datasets statistics. Topic has the following possible values: news articles, twitter messages, search queries, or sentences extracted from Wikipedia articles. Type of annotations can be NE (Named Entities i.e. persons, locations, organizations) and/or N (Nouns i.e. dataset annotates even simple nouns).
Exhaustive annotations means that these datasets annotate all entities that appear in their text and that fall into the selected categories of Type of Annotations. Recall 30/10 shows the percentage of the gold mentions for which their ground truth entity is among the first 30/10 candidate entities returned by our $p(e|m)$ dictionary. }

  \label{tab:datasets}

  \end{center}

\end{table*}

\begin{table*}[!h]
\scriptsize
\centering

\begin{tabular}{@{} l|c| c| c| c| c| c| c| c| c| c| c| c| c| c| @{}}
\multicolumn{1}{p{1cm}}{\textbf{F1@MA  F1@MI}}  &\rot{      AIDA A  } &\rot{      AIDA B  } &\rot{      MSNBC } &\rot{      OKE-2015} &\rot{OKE-2016} &\rot{      N3-Reuters-128   } &\rot{      N3-RSS-500 } &\rot{      Derczynski } &\rot{      KORE50 }\\
\hline
\hline
FREME& \multicolumn{1}{p{0.6cm}|}{23.9 38.3}& \multicolumn{1}{p{0.6cm}|}{24.3 37.0}& \multicolumn{1}{p{0.6cm}|}{17.6 22.6}& \multicolumn{1}{p{0.6cm}|}{26.8 32.7}& \multicolumn{1}{p{0.6cm}|}{23.0 29.0}& \multicolumn{1}{p{0.6cm}|}{28.2 33.5}& \multicolumn{1}{p{0.6cm}|}{34.6 32.4}& \multicolumn{1}{p{0.6cm}|}{31.8 20.4}& \multicolumn{1}{p{0.6cm}|}{13.3 15.5}\\
\hline
FOX& \multicolumn{1}{p{0.6cm}|}{54.9 58.1}& \multicolumn{1}{p{0.6cm}|}{58.4 57.4}& \multicolumn{1}{p{0.6cm}|}{26.3 21.2}& \multicolumn{1}{p{0.6cm}|}{56.4 58.7}& \multicolumn{1}{p{0.6cm}|}{50.1 51.4}& \multicolumn{1}{p{0.6cm}|}{53.8 55.2}& \multicolumn{1}{p{0.6cm}|}{37.7 37.2}& \multicolumn{1}{p{0.6cm}|}{43.4 41.5}& \multicolumn{1}{p{0.6cm}|}{30.1 32.9}\\
\hline
Babelfy& \multicolumn{1}{p{0.6cm}|}{42.1 48.1}& \multicolumn{1}{p{0.6cm}|}{43.3 49.5}& \multicolumn{1}{p{0.6cm}|}{40.2 44.9}& \multicolumn{1}{p{0.6cm}|}{42.8 45.7}& \multicolumn{1}{p{0.6cm}|}{40.1 39.7}& \multicolumn{1}{p{0.6cm}|}{29.8 33.6}& \multicolumn{1}{p{0.6cm}|}{36.6 34.6}& \multicolumn{1}{p{0.6cm}|}{32.3 34.3}& \multicolumn{1}{p{0.6cm}|}{\redc{53.0} \bluec{56.6}}\\
\hline
Entityclassifier.eu& \multicolumn{1}{p{0.6cm}|}{44.9 47.0}& \multicolumn{1}{p{0.6cm}|}{45.3 47.6}& \multicolumn{1}{p{0.6cm}|}{47.2 48.9}& \multicolumn{1}{p{0.6cm}|}{32.7 33.0}& \multicolumn{1}{p{0.6cm}|}{35.7 33.6}& \multicolumn{1}{p{0.6cm}|}{31.8 35.3}& \multicolumn{1}{p{0.6cm}|}{24.5 24.1}& \multicolumn{1}{p{0.6cm}|}{18.1 19.3}& \multicolumn{1}{p{0.6cm}|}{27.1 30.1}\\
\hline
Kea& \multicolumn{1}{p{0.6cm}|}{38.7 42.9}& \multicolumn{1}{p{0.6cm}|}{42.1 44.7}& \multicolumn{1}{p{0.6cm}|}{35.0 36.3}& \multicolumn{1}{p{0.6cm}|}{53.9 54.4}& \multicolumn{1}{p{0.6cm}|}{54.9 55.2}& \multicolumn{1}{p{0.6cm}|}{22.1 22.4}& \multicolumn{1}{p{0.6cm}|}{25.9 24.7}& \multicolumn{1}{p{0.6cm}|}{33.2 29.2}& \multicolumn{1}{p{0.6cm}|}{46.5 50.5}\\
\hline
DBpedia Spotlight& \multicolumn{1}{p{0.6cm}|}{55.5 58.3}& \multicolumn{1}{p{0.6cm}|}{53.1 58.8}& \multicolumn{1}{p{0.6cm}|}{49.6 49.2}& \multicolumn{1}{p{0.6cm}|}{49.6 52.2}& \multicolumn{1}{p{0.6cm}|}{49.2 49.8}& \multicolumn{1}{p{0.6cm}|}{36.0 38.0}& \multicolumn{1}{p{0.6cm}|}{35.5 35.1}& \multicolumn{1}{p{0.6cm}|}{44.0 41.5}& \multicolumn{1}{p{0.6cm}|}{28.8 35.4}\\
\hline
AIDA& \multicolumn{1}{p{0.6cm}|}{69.0 72.6}& \multicolumn{1}{p{0.6cm}|}{72.3 73.5}& \multicolumn{1}{p{0.6cm}|}{66.0 68.1}& \multicolumn{1}{p{0.6cm}|}{\bluec{61.7 65.8}}& \multicolumn{1}{p{0.6cm}|}{43.3 47.2}& \multicolumn{1}{p{0.6cm}|}{44.7 48.5}& \multicolumn{1}{p{0.6cm}|}{47.3 \redc{47.5}}& \multicolumn{1}{p{0.6cm}|}{41.5 34.8}& \multicolumn{1}{p{0.6cm}|}{\bluec{52.9} \redc{58.6}}\\
\hline
WAT& \multicolumn{1}{p{0.6cm}|}{69.7 73.3}& \multicolumn{1}{p{0.6cm}|}{71.6 73.6}& \multicolumn{1}{p{0.6cm}|}{66.7 68.9}& \multicolumn{1}{p{0.6cm}|}{58.4 60.7}& \multicolumn{1}{p{0.6cm}|}{\bluec{55.2 56.6}}& \multicolumn{1}{p{0.6cm}|}{\bluec{54.4} \redc{58.3}}& \multicolumn{1}{p{0.6cm}|}{47.6 \bluec{46.5}}& \multicolumn{1}{p{0.6cm}|}{47.1 43.6}& \multicolumn{1}{p{0.6cm}|}{39.9 52.2}\\
\hline
\begin{tabular}{@{}l@{}} \textbf{Best baseline} \end{tabular}& \multicolumn{1}{p{0.6cm}|}{\textbf{69.7 73.3}}& \multicolumn{1}{p{0.6cm}|}{\textbf{72.3 73.6}}& \multicolumn{1}{p{0.6cm}|}{\textbf{66.7 68.9}}& \multicolumn{1}{p{0.6cm}|}{\textbf{61.7 65.8}}& \multicolumn{1}{p{0.6cm}|}{\textbf{55.2 56.6}}& \multicolumn{1}{p{0.6cm}|}{\textbf{54.4 58.3}}& \multicolumn{1}{p{0.6cm}|}{\textbf{47.6 47.5}}& \multicolumn{1}{p{0.6cm}|}{\textbf{47.1 43.6}}& \multicolumn{1}{p{0.6cm}|}{\textbf{53.0 58.6}}\\
\hline
base model& \multicolumn{1}{p{0.6cm}|}{87.0 \bluec{89.5}}& \multicolumn{1}{p{0.6cm}|}{81.7 80.8}& \multicolumn{1}{p{0.6cm}|}{67.3 68.8}& \multicolumn{1}{p{0.6cm}|}{56.2 60.1}& \multicolumn{1}{p{0.6cm}|}{45.0 47.0}& \multicolumn{1}{p{0.6cm}|}{48.7 51.4}& \multicolumn{1}{p{0.6cm}|}{46.5 42.8}& \multicolumn{1}{p{0.6cm}|}{44.9 40.1}& \multicolumn{1}{p{0.6cm}|}{36.9 44.7}\\
\hline
base model + att& \multicolumn{1}{p{0.6cm}|}{\bluec{87.1} 89.3}& \multicolumn{1}{p{0.6cm}|}{\bluec{82.4 82.6}}& \multicolumn{1}{p{0.6cm}|}{72.6 72.6}& \multicolumn{1}{p{0.6cm}|}{59.0 63.0}& \multicolumn{1}{p{0.6cm}|}{49.2 51.6}& \multicolumn{1}{p{0.6cm}|}{51.0 54.7}& \multicolumn{1}{p{0.6cm}|}{\bluec{48.4} 44.6}& \multicolumn{1}{p{0.6cm}|}{\bluec{49.3 44.4}}& \multicolumn{1}{p{0.6cm}|}{38.0 45.1}\\
\hline
base model + att + global& \multicolumn{1}{p{0.6cm}|}{\redc{87.2 89.8}}& \multicolumn{1}{p{0.6cm}|}{\redc{83.2 82.8}}& \multicolumn{1}{p{0.6cm}|}{\redc{75.7} \bluec{74.7}}& \multicolumn{1}{p{0.6cm}|}{59.4 64.6}& \multicolumn{1}{p{0.6cm}|}{48.1 53.1}& \multicolumn{1}{p{0.6cm}|}{47.0 53.2}& \multicolumn{1}{p{0.6cm}|}{46.3 42.7}& \multicolumn{1}{p{0.6cm}|}{43.8 34.9}& \multicolumn{1}{p{0.6cm}|}{28.2 38.3}\\
\hline
\begin{tabular}{@{}c@{}} ED base model + att + global using \\ Stanford NER mentions\end{tabular} & \multicolumn{1}{p{0.6cm}|}{76.0 80.5}& \multicolumn{1}{p{0.6cm}|}{73.9 75.0}& \multicolumn{1}{p{0.6cm}|}{\bluec{74.7} \redc{74.8}}& \multicolumn{1}{p{0.6cm}|}{\redc{65.3} \redc{68.9}}& \multicolumn{1}{p{0.6cm}|}{\redc{57.7 59.3}}& \multicolumn{1}{p{0.6cm}|}{\redc{54.9} \bluec{56.9}}& \multicolumn{1}{p{0.6cm}|}{\redc{48.6} 46.1}& \multicolumn{1}{p{0.6cm}|}{\redc{50.2} \bluec{44.3}}& \multicolumn{1}{p{0.6cm}|}{41.9 47.8}\\
\hline
\end{tabular}
\caption{EL weak matching results on the Gerbil platform. Micro and Macro F1 scores are shown. We highlight in \redc{red} and \bluec{blue} the best and second best models, respectively. Training was done on AIDA-train set.}
\label{tab:el_weak}
\end{table*}

\begin{table*}[!h]
\scriptsize
\centering
\begin{adjustwidth}{-1cm}{}
\begin{tabular}{@{} l|c| c| c| c| c| c| c| c| c| c| c| c| c| c| c| @{}}
\multicolumn{1}{p{1cm}}{\textbf{F1@MA  F1@MI}}   &\rot{      ACE2004  } &\rot{      AIDA A  } &\rot{      AIDA B  } &\rot{      AQUAINT } &\rot{      MSNBC } &\rot{      DBpediaSpotlight } &\rot{      Derczynski } &\rot{      ERD2014 } &\rot{      GERDAQ-Dev } &\rot{      GERDAQ-Test } &\rot{      KORE50 } &\rot{      N3-Reuters-128   } &\rot{      N3-RSS-500 } &\rot{      OKE-2015} &\rot{OKE-2016}\\
\hline
\hline
AGDISTIS& \multicolumn{1}{p{0.6cm}|}{78.1 65.6}& \multicolumn{1}{p{0.6cm}|}{49.2 55.6}& \multicolumn{1}{p{0.6cm}|}{54.2 54.9}& \multicolumn{1}{p{0.6cm}|}{58.9 60.4}& \multicolumn{1}{p{0.6cm}|}{72.9 74.0}& \multicolumn{1}{p{0.6cm}|}{35.3 39.3}& \multicolumn{1}{p{0.6cm}|}{50.8 44.5}& \multicolumn{1}{p{0.6cm}|}{63.2 32.0}& \multicolumn{1}{p{0.6cm}|}{21.4 17.8}& \multicolumn{1}{p{0.6cm}|}{22.8 20.3}& \multicolumn{1}{p{0.6cm}|}{30.0 33.3}& \multicolumn{1}{p{0.6cm}|}{\bluec{68.5} 63.8}& \multicolumn{1}{p{0.6cm}|}{51.4 51.8}& \multicolumn{1}{p{0.6cm}|}{61.4 62.0}& \multicolumn{1}{p{0.6cm}|}{59.6 58.8}\\
\hline
FREME& \multicolumn{1}{p{0.6cm}|}{73.5 60.3}& \multicolumn{1}{p{0.6cm}|}{25.3 33.0}& \multicolumn{1}{p{0.6cm}|}{25.7 40.3}& \multicolumn{1}{p{0.6cm}|}{43.7 57.6}& \multicolumn{1}{p{0.6cm}|}{17.9 24.3}& \multicolumn{1}{p{0.6cm}|}{37.9 52.5}& \multicolumn{1}{p{0.6cm}|}{48.3 38.2}& \multicolumn{1}{p{0.6cm}|}{71.3 51.8}& \multicolumn{1}{p{0.6cm}|}{37.6 44.2}& \multicolumn{1}{p{0.6cm}|}{33.4 39.6}& \multicolumn{1}{p{0.6cm}|}{13.8 17.4}& \multicolumn{1}{p{0.6cm}|}{30.2 31.7}& \multicolumn{1}{p{0.6cm}|}{45.5 43.9}& \multicolumn{1}{p{0.6cm}|}{25.0 33.3}& \multicolumn{1}{p{0.6cm}|}{27.4 35.0}\\
\hline
FOX& \multicolumn{1}{p{0.6cm}|}{38.6 0.0}& \multicolumn{1}{p{0.6cm}|}{55.5 58.9}& \multicolumn{1}{p{0.6cm}|}{59.6 58.0}& \multicolumn{1}{p{0.6cm}|}{0.0 0.0}& \multicolumn{1}{p{0.6cm}|}{16.2 13.3}& \multicolumn{1}{p{0.6cm}|}{11.7 15.9}& \multicolumn{1}{p{0.6cm}|}{51.2 44.3}& \multicolumn{1}{p{0.6cm}|}{51.7 0.0}& \multicolumn{1}{p{0.6cm}|}{11.2 0.0}& \multicolumn{1}{p{0.6cm}|}{10.6 0.0}& \multicolumn{1}{p{0.6cm}|}{26.9 31.1}& \multicolumn{1}{p{0.6cm}|}{56.6 58.1}& \multicolumn{1}{p{0.6cm}|}{52.9 53.6}& \multicolumn{1}{p{0.6cm}|}{54.9 59.0}& \multicolumn{1}{p{0.6cm}|}{50.1 51.4}\\
\hline
Babelfy& \multicolumn{1}{p{0.6cm}|}{76.6 61.9}& \multicolumn{1}{p{0.6cm}|}{66.4 71.9}& \multicolumn{1}{p{0.6cm}|}{70.0 75.8}& \multicolumn{1}{p{0.6cm}|}{70.3 70.4}& \multicolumn{1}{p{0.6cm}|}{72.7 76.2}& \multicolumn{1}{p{0.6cm}|}{51.0 51.8}& \multicolumn{1}{p{0.6cm}|}{64.4 63.2}& \multicolumn{1}{p{0.6cm}|}{69.6 45.0}& \multicolumn{1}{p{0.6cm}|}{30.6 34.3}& \multicolumn{1}{p{0.6cm}|}{31.5 35.0}& \multicolumn{1}{p{0.6cm}|}{\redc{70.5 74.1}}& \multicolumn{1}{p{0.6cm}|}{54.6 55.4}& \multicolumn{1}{p{0.6cm}|}{62.8 64.1}& \multicolumn{1}{p{0.6cm}|}{65.1 65.9}& \multicolumn{1}{p{0.6cm}|}{60.2 59.8}\\
\hline
Entityclassifier.eu& \multicolumn{1}{p{0.6cm}|}{72.6 61.9}& \multicolumn{1}{p{0.6cm}|}{53.6 53.7}& \multicolumn{1}{p{0.6cm}|}{54.3 53.9}& \multicolumn{1}{p{0.6cm}|}{39.9 43.1}& \multicolumn{1}{p{0.6cm}|}{54.1 54.6}& \multicolumn{1}{p{0.6cm}|}{19.9 25.5}& \multicolumn{1}{p{0.6cm}|}{43.2 32.6}& \multicolumn{1}{p{0.6cm}|}{51.7 0.0}& \multicolumn{1}{p{0.6cm}|}{11.2 0.0}& \multicolumn{1}{p{0.6cm}|}{10.6 0.0}& \multicolumn{1}{p{0.6cm}|}{25.9 29.9}& \multicolumn{1}{p{0.6cm}|}{40.3 44.4}& \multicolumn{1}{p{0.6cm}|}{37.6 44.9}& \multicolumn{1}{p{0.6cm}|}{35.3 37.6}& \multicolumn{1}{p{0.6cm}|}{40.3 41.4}\\
\hline
Kea& \multicolumn{1}{p{0.6cm}|}{87.4 79.5}& \multicolumn{1}{p{0.6cm}|}{62.6 63.5}& \multicolumn{1}{p{0.6cm}|}{69.0 68.5}& \multicolumn{1}{p{0.6cm}|}{79.9 80.6}& \multicolumn{1}{p{0.6cm}|}{80.9 79.3}& \multicolumn{1}{p{0.6cm}|}{\bluec{75.5} 72.2}& \multicolumn{1}{p{0.6cm}|}{63.6 61.5}& \multicolumn{1}{p{0.6cm}|}{\bluec{87.4} 71.0}& \multicolumn{1}{p{0.6cm}|}{56.8 60.6}& \multicolumn{1}{p{0.6cm}|}{\bluec{59.7 63.3}}& \multicolumn{1}{p{0.6cm}|}{50.9 60.8}& \multicolumn{1}{p{0.6cm}|}{62.7 62.7}& \multicolumn{1}{p{0.6cm}|}{58.6 64.5}& \multicolumn{1}{p{0.6cm}|}{68.5 69.1}& \multicolumn{1}{p{0.6cm}|}{67.7 69.1}\\
\hline
DBpedia Spotlight& \multicolumn{1}{p{0.6cm}|}{73.4 53.9}& \multicolumn{1}{p{0.6cm}|}{51.5 53.2}& \multicolumn{1}{p{0.6cm}|}{53.7 56.1}& \multicolumn{1}{p{0.6cm}|}{50.6 51.8}& \multicolumn{1}{p{0.6cm}|}{43.6 42.1}& \multicolumn{1}{p{0.6cm}|}{70.1 72.6}& \multicolumn{1}{p{0.6cm}|}{50.3 43.3}& \multicolumn{1}{p{0.6cm}|}{65.2 38.4}& \multicolumn{1}{p{0.6cm}|}{49.2 54.4}& \multicolumn{1}{p{0.6cm}|}{46.8 54.0}& \multicolumn{1}{p{0.6cm}|}{48.7 52.3}& \multicolumn{1}{p{0.6cm}|}{41.8 43.4}& \multicolumn{1}{p{0.6cm}|}{42.6 34.6}& \multicolumn{1}{p{0.6cm}|}{30.4 35.8}& \multicolumn{1}{p{0.6cm}|}{43.0 43.1}\\
\hline
AIDA& \multicolumn{1}{p{0.6cm}|}{89.5 79.8}& \multicolumn{1}{p{0.6cm}|}{71.4 74.7}& \multicolumn{1}{p{0.6cm}|}{77.8 77.0}& \multicolumn{1}{p{0.6cm}|}{56.6 57.1}& \multicolumn{1}{p{0.6cm}|}{70.3 74.6}& \multicolumn{1}{p{0.6cm}|}{21.4 24.9}& \multicolumn{1}{p{0.6cm}|}{64.5 63.9}& \multicolumn{1}{p{0.6cm}|}{79.1 63.7}& \multicolumn{1}{p{0.6cm}|}{26.5 23.5}& \multicolumn{1}{p{0.6cm}|}{27.2 25.0}& \multicolumn{1}{p{0.6cm}|}{\bluec{63.2 69.1}}& \multicolumn{1}{p{0.6cm}|}{51.2 56.9}& \multicolumn{1}{p{0.6cm}|}{62.9 65.5}& \multicolumn{1}{p{0.6cm}|}{61.6 63.0}& \multicolumn{1}{p{0.6cm}|}{0.0 0.0}\\
\hline
WAT& \multicolumn{1}{p{0.6cm}|}{86.9 79.6}& \multicolumn{1}{p{0.6cm}|}{75.6 78.5}& \multicolumn{1}{p{0.6cm}|}{79.8 80.5}& \multicolumn{1}{p{0.6cm}|}{75.6 75.4}& \multicolumn{1}{p{0.6cm}|}{79.7 78.8}& \multicolumn{1}{p{0.6cm}|}{68.8 67.1}& \multicolumn{1}{p{0.6cm}|}{\redc{70.4 69.5}}& \multicolumn{1}{p{0.6cm}|}{86.5 \bluec{76.0}}& \multicolumn{1}{p{0.6cm}|}{\bluec{61.2 63.9}}& \multicolumn{1}{p{0.6cm}|}{57.4 60.7}& \multicolumn{1}{p{0.6cm}|}{52.4 62.2}& \multicolumn{1}{p{0.6cm}|}{59.2 63.1}& \multicolumn{1}{p{0.6cm}|}{62.8 63.9}& \multicolumn{1}{p{0.6cm}|}{62.2 64.9}& \multicolumn{1}{p{0.6cm}|}{0.0 0.0}\\
\hline
PBOH& \multicolumn{1}{p{0.6cm}|}{88.0 83.2}& \multicolumn{1}{p{0.6cm}|}{77.3 80.1}& \multicolumn{1}{p{0.6cm}|}{79.0 80.4}& \multicolumn{1}{p{0.6cm}|}{\redc{83.3 84.1}}& \multicolumn{1}{p{0.6cm}|}{85.5 86.1}& \multicolumn{1}{p{0.6cm}|}{\redc{80.5 80.1}}& \multicolumn{1}{p{0.6cm}|}{63.6 64.8}& \multicolumn{1}{p{0.6cm}|}{\redc{91.0 80.7}}& \multicolumn{1}{p{0.6cm}|}{\redc{67.0 68.8}}& \multicolumn{1}{p{0.6cm}|}{\redc{64.4 67.8}}& \multicolumn{1}{p{0.6cm}|}{58.6 63.2}& \multicolumn{1}{p{0.6cm}|}{\redc{72.1 69.0}}& \multicolumn{1}{p{0.6cm}|}{47.6 55.2}& \multicolumn{1}{p{0.6cm}|}{67.4 67.6}& \multicolumn{1}{p{0.6cm}|}{0.0 0.0}\\
\hline
\begin{tabular}{@{}l@{}} \textbf{Best baseline} \end{tabular}& \multicolumn{1}{p{0.6cm}|}{\textbf{89.5 83.2}}& \multicolumn{1}{p{0.6cm}|}{\textbf{77.3 80.1}}& \multicolumn{1}{p{0.6cm}|}{\textbf{79.8 80.5}}& \multicolumn{1}{p{0.6cm}|}{\textbf{83.3 84.1}}& \multicolumn{1}{p{0.6cm}|}{\textbf{85.5 86.1}}& \multicolumn{1}{p{0.6cm}|}{\textbf{80.5 80.1}}& \multicolumn{1}{p{0.6cm}|}{\textbf{70.4 69.5}}& \multicolumn{1}{p{0.6cm}|}{\textbf{91.0 80.7}}& \multicolumn{1}{p{0.6cm}|}{\textbf{67.0 68.8}}& \multicolumn{1}{p{0.6cm}|}{\textbf{64.4 67.8}}& \multicolumn{1}{p{0.6cm}|}{\textbf{70.5 74.1}}& \multicolumn{1}{p{0.6cm}|}{\textbf{72.1 69.0}}& \multicolumn{1}{p{0.6cm}|}{\textbf{62.9 65.5}}& \multicolumn{1}{p{0.6cm}|}{\textbf{68.5 69.1}}& \multicolumn{1}{p{0.6cm}|}{\textbf{67.7 69.1}}\\
\hline
ed base model& \multicolumn{1}{p{0.6cm}|}{\bluec{89.9} 83.5}& \multicolumn{1}{p{0.6cm}|}{\bluec{86.5} 88.8}& \multicolumn{1}{p{0.6cm}|}{82.7 82.7}& \multicolumn{1}{p{0.6cm}|}{78.4 79.7}& \multicolumn{1}{p{0.6cm}|}{82.4 82.5}& \multicolumn{1}{p{0.6cm}|}{71.1 71.0}& \multicolumn{1}{p{0.6cm}|}{63.5 62.7}& \multicolumn{1}{p{0.6cm}|}{81.0 65.3}& \multicolumn{1}{p{0.6cm}|}{54.2 58.4}& \multicolumn{1}{p{0.6cm}|}{52.7 57.6}& \multicolumn{1}{p{0.6cm}|}{45.1 50.8}& \multicolumn{1}{p{0.6cm}|}{62.9 66.4}& \multicolumn{1}{p{0.6cm}|}{64.0 65.8}& \multicolumn{1}{p{0.6cm}|}{69.7 70.0}& \multicolumn{1}{p{0.6cm}|}{71.9 73.0}\\
\hline
ed base model + att& \multicolumn{1}{p{0.6cm}|}{89.5 \bluec{84.1}}& \multicolumn{1}{p{0.6cm}|}{86.0 \bluec{88.9}}& \multicolumn{1}{p{0.6cm}|}{\bluec{83.6} \redc{83.1}}& \multicolumn{1}{p{0.6cm}|}{81.2 82.2}& \multicolumn{1}{p{0.6cm}|}{\bluec{88.4} \redc{86.4}}& \multicolumn{1}{p{0.6cm}|}{67.8 72.8}& \multicolumn{1}{p{0.6cm}|}{64.3 65.0}& \multicolumn{1}{p{0.6cm}|}{79.9 65.2}& \multicolumn{1}{p{0.6cm}|}{41.8 49.2}& \multicolumn{1}{p{0.6cm}|}{46.6 54.7}& \multicolumn{1}{p{0.6cm}|}{48.3 55.5}& \multicolumn{1}{p{0.6cm}|}{62.5 \bluec{67.3}}& \multicolumn{1}{p{0.6cm}|}{\redc{67.2} \bluec{68.4}}& \multicolumn{1}{p{0.6cm}|}{\redc{73.7 74.5}}& \multicolumn{1}{p{0.6cm}|}{\bluec{75.1 76.1}}\\
\hline
ed base model + att + global& \multicolumn{1}{p{0.6cm}|}{\redc{92.0 85.5}}& \multicolumn{1}{p{0.6cm}|}{\redc{86.6 89.0}}& \multicolumn{1}{p{0.6cm}|}{\redc{83.8} \bluec{83.0}}& \multicolumn{1}{p{0.6cm}|}{\bluec{82.3 83.2}}& \multicolumn{1}{p{0.6cm}|}{\redc{88.5} \bluec{86.2}}& \multicolumn{1}{p{0.6cm}|}{74.4 \bluec{78.8}}& \multicolumn{1}{p{0.6cm}|}{\bluec{65.3 65.4}}& \multicolumn{1}{p{0.6cm}|}{76.9 61.4}& \multicolumn{1}{p{0.6cm}|}{38.6 46.3}& \multicolumn{1}{p{0.6cm}|}{43.2 53.7}& \multicolumn{1}{p{0.6cm}|}{52.4 60.8}& \multicolumn{1}{p{0.6cm}|}{63.4 \bluec{67.3}}& \multicolumn{1}{p{0.6cm}|}{\bluec{66.6} \redc{68.6}}& \multicolumn{1}{p{0.6cm}|}{\bluec{73.2 74.0}}& \multicolumn{1}{p{0.6cm}|}{\redc{76.7 78.1}}\\
\hline
\end{tabular}
\caption{ED results on the Gerbil platform. Micro and Macro F1 scores are shown. We highlight in \redc{red} and \bluec{blue} the best and second best models, respectively. Training was done on AIDA-train set.}
\label{tab:ed}
\end{adjustwidth}
\end{table*}


\begin{table*}[!h]
\scriptsize
\centering
\begin{tabular}{@{} l|c| c| c| c| c| c|@{}}
\multicolumn{1}{p{1cm}}{\textbf{F1@MA  F1@MI}}   &\rot{AIDA A} &\rot{AIDA B} &\rot{ACE2004} &\rot{AQUAINT} &\rot{MSNBC} \\
\hline
\hline
ed base model& \multicolumn{1}{p{0.6cm}|}{92.1 93.3}& \multicolumn{1}{p{0.6cm}|}{88.4 86.7}& \multicolumn{1}{p{0.6cm}|}{84.4 85.4}& \multicolumn{1}{p{0.6cm}|}{85.9 86.1}& \multicolumn{1}{p{0.6cm}|}{88.2 89.6}\\
\hline
ed base model + att& \multicolumn{1}{p{0.6cm}|}{91.7 93.5}& \multicolumn{1}{p{0.6cm}|}{89.0 87.2}& \multicolumn{1}{p{0.6cm}|}{87.3 87.0}& \multicolumn{1}{p{0.6cm}|}{89.1 89.4}& \multicolumn{1}{p{0.6cm}|}{93.2 92.3} \\
\hline
ed base model + att + global& \multicolumn{1}{p{0.6cm}|}{92.4 93.7}& \multicolumn{1}{p{0.6cm}|}{89.1 87.3}& \multicolumn{1}{p{0.6cm}|}{89.2 88.1}& \multicolumn{1}{p{0.6cm}|}{89.4 89.7}& \multicolumn{1}{p{0.6cm}|}{93.3 92.5} \\
\hline
\end{tabular}
\caption{\textbf{ED} results matched locally (not on Gerbil) using KB entity IDs. Micro and Macro F1 scores are shown. For the ED task there is a significant discrepancy between the local and the Gerbil scores which likely has two main causes: a different URL-based matching scheme used in Gerbil\footnotemark and parsing or alignment errors.}
\label{tab:ed_local_results}
\end{table*}
\footnotetext{\url{https://github.com/dice-group/gerbil/issues/98}}

\begin{table*}[!h]
\scriptsize
\centering
\begin{tabular}{@{} l|c| c| @{}} 
\multicolumn{1}{p{1cm}}{\textbf{Rec@MA  Rec@MI}} &\rot{      ACE2004  } &\rot{      AQUAINT }\\
\hline
\hline
FREME& \multicolumn{1}{p{0.6cm}|}{\redc{44.8} 30.8}& \multicolumn{1}{p{0.6cm}|}{23.7 23.7}\\
\hline
FOX& \multicolumn{1}{p{0.6cm}|}{0.0 0.0}& \multicolumn{1}{p{0.6cm}|}{0.0 0.0}\\
\hline
Babelfy& \multicolumn{1}{p{0.6cm}|}{10.0 17.8}& \multicolumn{1}{p{0.6cm}|}{34.9 35.8}\\
\hline
Entityclassifier.eu& \multicolumn{1}{p{0.6cm}|}{35.7 59.7}& \multicolumn{1}{p{0.6cm}|}{28.4 29.2}\\
\hline
Kea& \multicolumn{1}{p{0.6cm}|}{23.7 40.3}& \multicolumn{1}{p{0.6cm}|}{36.1 35.9}\\
\hline
DBpedia Spotlight& \multicolumn{1}{p{0.6cm}|}{39.1 \bluec{60.5}}& \multicolumn{1}{p{0.6cm}|}{\redc{45.1 45.2}}\\
\hline
AIDA& \multicolumn{1}{p{0.6cm}|}{28.0 48.6}& \multicolumn{1}{p{0.6cm}|}{36.0 36.5}\\
\hline
WAT& \multicolumn{1}{p{0.6cm}|}{26.8 44.7}& \multicolumn{1}{p{0.6cm}|}{35.1 35.5}\\
\hline
\begin{tabular}{@{}l@{}} \textbf{Best baseline} \end{tabular}& \multicolumn{1}{p{0.6cm}|}{\textbf{44.8 60.5}}& \multicolumn{1}{p{0.6cm}|}{\textbf{45.1 45.2}}\\
\hline
model+att+global& \multicolumn{1}{p{0.6cm}|}{\bluec{40.3} \redc{68.3}}& \multicolumn{1}{p{0.6cm}|}{\bluec{39.4 40.4}}\\
\hline  
\end{tabular}
\caption{El strong matching results on the Gerbil platform. Micro and Macro recall scores are shown for ACE2004 and AQUAINT datasets that only annotate each entity once, thus F1 scores not being useful. We highlight in \redc{red} and \bluec{blue} the best and second best models, respectively. Our models were trained on AIDA as before.}
\end{table*}

\begin{table*}[!h]
\scriptsize
\centering

\begin{tabular}{@{} l|c| c| c| @{}}
\multicolumn{1}{p{1cm}}{\textbf{F1@MA  F1@MI}}   &\rot{      GERDAQ-Dev } &\rot{      GERDAQ-Test } &\rot{      DBpediaSpotlight }\\
\hline
\hline
FREME& \multicolumn{1}{p{0.6cm}|}{11.2 0.0}& \multicolumn{1}{p{0.6cm}|}{10.6 0.0}& \multicolumn{1}{p{0.6cm}|}{11.6 13.0}\\
\hline
FOX& \multicolumn{1}{p{0.6cm}|}{11.2 0.0}& \multicolumn{1}{p{0.6cm}|}{10.6 0.0}& \multicolumn{1}{p{0.6cm}|}{11.3 15.3}\\
\hline
Babelfy& \multicolumn{1}{p{0.6cm}|}{19.6 15.5}& \multicolumn{1}{p{0.6cm}|}{19.6 17.5}& \multicolumn{1}{p{0.6cm}|}{8.7 13.1}\\
\hline
Entityclassifier.eu& \multicolumn{1}{p{0.6cm}|}{11.2 0.0}& \multicolumn{1}{p{0.6cm}|}{10.6 0.0}& \multicolumn{1}{p{0.6cm}|}{18.3 23.1}\\
\hline
Kea& \multicolumn{1}{p{0.6cm}|}{\bluec{40.2 40.2}}& \multicolumn{1}{p{0.6cm}|}{\redc{40.2 43.7}}& \multicolumn{1}{p{0.6cm}|}{\redc{42.0 42.6}}\\
\hline
DBpedia Spotlight& \multicolumn{1}{p{0.6cm}|}{29.6 27.8}& \multicolumn{1}{p{0.6cm}|}{29.4 31.4}& \multicolumn{1}{p{0.6cm}|}{31.8 \bluec{37.6}}\\
\hline
AIDA& \multicolumn{1}{p{0.6cm}|}{11.2 0.0}& \multicolumn{1}{p{0.6cm}|}{10.6 0.0}& \multicolumn{1}{p{0.6cm}|}{14.4 19.1}\\
\hline
WAT& \multicolumn{1}{p{0.6cm}|}{12.9 3.3}& \multicolumn{1}{p{0.6cm}|}{12.6 6.0}& \multicolumn{1}{p{0.6cm}|}{14.5 17.0}\\
\hline
\begin{tabular}{@{}l@{}} \textbf{Best baseline} \end{tabular}& \multicolumn{1}{p{0.6cm}|}{\textbf{40.2 40.2}}& \multicolumn{1}{p{0.6cm}|}{\textbf{40.2 43.7}}& \multicolumn{1}{p{0.6cm}|}{\textbf{42.0 42.6}}\\
\hline
model+att+global on gerdaq train& \multicolumn{1}{p{0.6cm}|}{\redc{42.7 44.9}}& \multicolumn{1}{p{0.6cm}|}{\bluec{40.0 41.2}}& \multicolumn{1}{p{0.6cm}|}{\bluec{33.4} 34.8}\\
\hline
\end{tabular}
\caption{El strong matching results on the Gerbil platform. Micro and Macro F1 scores are shown. Training was done on Gerdaq train dataset. We highlight in \redc{red} and \bluec{blue} the best and second best models, respectively.}
\label{tab:gerdaq_results}
\end{table*}

\begin{table*}[!h]
\scriptsize
\centering

\begin{tabular}{@{} l|c| c| @{}}
\multicolumn{1}{p{1cm}}{\textbf{F1@MA  F1@MI}}   &\rot{MicropDev} &\rot{MicropTest}\\
\hline
\hline
FREME& \multicolumn{1}{p{0.6cm}|}{3.9 3.8}& \multicolumn{1}{p{0.6cm}|}{\redc{77.9} 3.7}\\
\hline
FOX& \multicolumn{1}{p{0.6cm}|}{6.1 7.5}& \multicolumn{1}{p{0.6cm}|}{71.9 3.7}\\
\hline
Babelfy& \multicolumn{1}{p{0.6cm}|}{6.2 9.5}& \multicolumn{1}{p{0.6cm}|}{52.5 2.8}\\
\hline
Entityclassifier.eu& \multicolumn{1}{p{0.6cm}|}{23.5 29.1}& \multicolumn{1}{p{0.6cm}|}{9.2 2.4}\\
\hline
Kea& \multicolumn{1}{p{0.6cm}|}{\bluec{37.3 38.8}}& \multicolumn{1}{p{0.6cm}|}{51.6 6.5}\\
\hline
DBpedia Spotlight& \multicolumn{1}{p{0.6cm}|}{29.1 34.0}& \multicolumn{1}{p{0.6cm}|}{13.6 5.2}\\
\hline
AIDA& \multicolumn{1}{p{0.6cm}|}{6.4 9.7}& \multicolumn{1}{p{0.6cm}|}{73.3 5.0}\\
\hline
WAT& \multicolumn{1}{p{0.6cm}|}{0.0 0.0}& \multicolumn{1}{p{0.6cm}|}{0.0 0.0}\\
\hline
\begin{tabular}{@{}l@{}} \textbf{Best baseline} \end{tabular}& \multicolumn{1}{p{0.6cm}|}{\textbf{37.3 38.8}}& \multicolumn{1}{p{0.6cm}|}{\textbf{77.9 6.5}}\\
\hline
model on micropost train& \multicolumn{1}{p{0.6cm}|}{\redc{57.3 63.7}}& \multicolumn{1}{p{0.6cm}|}{17.4 \bluec{10.0}}\\
\hline
\end{tabular}
\caption{El strong matching results on the Gerbil platform when training on Micropost-train 2016. Micro and Macro F1 scores are shown. The majority of the errors are due to the newer KB version used in this dataset. We highlight in \redc{red} and \bluec{blue} the best and second best models, respectively.}
\label{tab:micropost_results}
\end{table*}

%% file: conll2018.bbl
\begin{thebibliography}{37}
\expandafter\ifx\csname natexlab\endcsname\relax\def\natexlab#1{#1}\fi

\bibitem[{Carmel et~al.(2014)Carmel, Chang, Gabrilovich, Hsu, and
  Wang}]{carmel2014erd}
David Carmel, Ming-Wei Chang, Evgeniy Gabrilovich, Bo-June~Paul Hsu, and
  Kuansan Wang. 2014.
\newblock Erd'14: entity recognition and disambiguation challenge.
\newblock In \emph{ACM SIGIR Forum}, volume~48, pages 63--77. ACM.

\bibitem[{Ceccarelli et~al.(2013)Ceccarelli, Lucchese, Orlando, Perego, and
  Trani}]{ceccarelli2013dexter}
Diego Ceccarelli, Claudio Lucchese, Salvatore Orlando, Raffaele Perego, and
  Salvatore Trani. 2013.
\newblock Dexter: an open source framework for entity linking.
\newblock In \emph{Proceedings of the sixth international workshop on
  Exploiting semantic annotations in information retrieval}, pages 17--20. ACM.

\bibitem[{Chiu and Nichols(2016)}]{chiu2016named}
Jason~PC Chiu and Eric Nichols. 2016.
\newblock Named entity recognition with bidirectional lstm-cnns.
\newblock \emph{Transactions of the Association for Computational Linguistics},
  4:357--370.

\bibitem[{Cornolti et~al.(2016)Cornolti, Ferragina, Ciaramita, R{\"u}d, and
  Sch{\"u}tze}]{cornolti2016piggyback}
Marco Cornolti, Paolo Ferragina, Massimiliano Ciaramita, Stefan R{\"u}d, and
  Hinrich Sch{\"u}tze. 2016.
\newblock A piggyback system for joint entity mention detection and linking in
  web queries.
\newblock In \emph{Proceedings of the 25th International Conference on World
  Wide Web}, pages 567--578. International World Wide Web Conferences Steering
  Committee.

\bibitem[{Daiber et~al.(2013)Daiber, Jakob, Hokamp, and
  Mendes}]{daiber2013improving}
Joachim Daiber, Max Jakob, Chris Hokamp, and Pablo~N Mendes. 2013.
\newblock Improving efficiency and accuracy in multilingual entity extraction.
\newblock In \emph{Proceedings of the 9th International Conference on Semantic
  Systems}, pages 121--124. ACM.

\bibitem[{Derczynski et~al.(2015)Derczynski, Maynard, Rizzo, van Erp, Gorrell,
  Troncy, Petrak, and Bontcheva}]{derczynski2015analysis}
Leon Derczynski, Diana Maynard, Giuseppe Rizzo, Marieke van Erp, Genevieve
  Gorrell, Rapha{\"e}l Troncy, Johann Petrak, and Kalina Bontcheva. 2015.
\newblock Analysis of named entity recognition and linking for tweets.
\newblock \emph{Information Processing \& Management}, 51(2):32--49.

\bibitem[{van Erp et~al.(2013)van Erp, Rizzo, and Troncy}]{van2013learning}
MGJ van Erp, G~Rizzo, and R~Troncy. 2013.
\newblock Learning with the web: Spotting named entities on the intersection of
  nerd and machine learning.
\newblock In \emph{CEUR workshop proceedings}, pages 27--30.

\bibitem[{Finkel et~al.(2005)Finkel, Grenager, and
  Manning}]{finkel2005incorporating}
Jenny~Rose Finkel, Trond Grenager, and Christopher Manning. 2005.
\newblock Incorporating non-local information into information extraction
  systems by gibbs sampling.
\newblock In \emph{Proceedings of the 43rd annual meeting on association for
  computational linguistics}, pages 363--370. Association for Computational
  Linguistics.

\bibitem[{Ganea et~al.(2016)Ganea, Ganea, Lucchi, Eickhoff, and
  Hofmann}]{ganea2016probabilistic}
Octavian-Eugen Ganea, Marina Ganea, Aurelien Lucchi, Carsten Eickhoff, and
  Thomas Hofmann. 2016.
\newblock Probabilistic bag-of-hyperlinks model for entity linking.
\newblock In \emph{Proceedings of the 25th International Conference on World
  Wide Web}, pages 927--938. International World Wide Web Conferences Steering
  Committee.

\bibitem[{Ganea and Hofmann(2017)}]{ganea2017deep}
Octavian-Eugen Ganea and Thomas Hofmann. 2017.
\newblock Deep joint entity disambiguation with local neural attention.
\newblock In \emph{Proceedings of the 2017 Conference on Empirical Methods in
  Natural Language Processing}, pages 2619--2629.

\bibitem[{He et~al.(2013)He, Liu, Li, Zhou, Zhang, and Wang}]{he2013learning}
Zhengyan He, Shujie Liu, Mu~Li, Ming Zhou, Longkai Zhang, and Houfeng Wang.
  2013.
\newblock Learning entity representation for entity disambiguation.
\newblock In \emph{Proceedings of the 51st Annual Meeting of the Association
  for Computational Linguistics (Volume 2: Short Papers)}, volume~2, pages
  30--34.

\bibitem[{Hochreiter and Schmidhuber(1997)}]{hochreiter1997long}
Sepp Hochreiter and J{\"u}rgen Schmidhuber. 1997.
\newblock Long short-term memory.
\newblock \emph{Neural computation}, 9(8):1735--1780.

\bibitem[{Hoffart et~al.(2012)Hoffart, Seufert, Nguyen, Theobald, and
  Weikum}]{hoffart2012kore}
Johannes Hoffart, Stephan Seufert, Dat~Ba Nguyen, Martin Theobald, and Gerhard
  Weikum. 2012.
\newblock Kore: keyphrase overlap relatedness for entity disambiguation.
\newblock In \emph{Proceedings of the 21st ACM international conference on
  Information and knowledge management}, pages 545--554. ACM.

\bibitem[{Hoffart et~al.(2011)Hoffart, Yosef, Bordino, F{\"u}rstenau, Pinkal,
  Spaniol, Taneva, Thater, and Weikum}]{hoffart2011robust}
Johannes Hoffart, Mohamed~Amir Yosef, Ilaria Bordino, Hagen F{\"u}rstenau,
  Manfred Pinkal, Marc Spaniol, Bilyana Taneva, Stefan Thater, and Gerhard
  Weikum. 2011.
\newblock Robust disambiguation of named entities in text.
\newblock In \emph{Proceedings of the Conference on Empirical Methods in
  Natural Language Processing}, pages 782--792. Association for Computational
  Linguistics.

\bibitem[{Ji et~al.(2017)Ji, Tan, Martschat, Choi, and Smith}]{ji2017dynamic}
Yangfeng Ji, Chenhao Tan, Sebastian Martschat, Yejin Choi, and Noah~A Smith.
  2017.
\newblock Dynamic entity representations in neural language models.
\newblock In \emph{Proceedings of the 2017 Conference on Empirical Methods in
  Natural Language Processing}, pages 1830--1839.

\bibitem[{Kingma and Ba(2014)}]{kingma2014adam}
Diederik~P Kingma and Jimmy Ba. 2014.
\newblock Adam: A method for stochastic optimization.
\newblock \emph{arXiv preprint arXiv:1412.6980}.

\bibitem[{Lample et~al.(2016)Lample, Ballesteros, Subramanian, Kawakami, and
  Dyer}]{lample2016neural}
Guillaume Lample, Miguel Ballesteros, Sandeep Subramanian, Kazuya Kawakami, and
  Chris Dyer. 2016.
\newblock Neural architectures for named entity recognition.
\newblock In \emph{Proceedings of NAACL-HLT}, pages 260--270.

\bibitem[{Le and Titov(2018)}]{le2018improving}
Phong Le and Ivan Titov. 2018.
\newblock Improving entity linking by modeling latent relations between
  mentions.
\newblock \emph{arXiv preprint arXiv:1804.10637}.

\bibitem[{Lee et~al.(2017)Lee, He, Lewis, and Zettlemoyer}]{lee2017end}
Kenton Lee, Luheng He, Mike Lewis, and Luke Zettlemoyer. 2017.
\newblock End-to-end neural coreference resolution.
\newblock In \emph{Proceedings of the 2017 Conference on Empirical Methods in
  Natural Language Processing}, pages 188--197.

\bibitem[{Liu et~al.(2017)Liu, Shang, Xu, Ren, Gui, Peng, and
  Han}]{liu2017empower}
Liyuan Liu, Jingbo Shang, Frank Xu, Xiang Ren, Huan Gui, Jian Peng, and Jiawei
  Han. 2017.
\newblock Empower sequence labeling with task-aware neural language model.
\newblock \emph{arXiv preprint arXiv:1709.04109}.

\bibitem[{Luo et~al.(2015)Luo, Huang, Lin, and Nie}]{luo2015joint}
Gang Luo, Xiaojiang Huang, Chin-Yew Lin, and Zaiqing Nie. 2015.
\newblock Joint entity recognition and disambiguation.
\newblock In \emph{Proceedings of the 2015 Conference on Empirical Methods in
  Natural Language Processing}, pages 879--888.

\bibitem[{Mendes et~al.(2011)Mendes, Jakob, Garc{\'\i}a-Silva, and
  Bizer}]{mendes2011dbpedia}
Pablo~N Mendes, Max Jakob, Andr{\'e}s Garc{\'\i}a-Silva, and Christian Bizer.
  2011.
\newblock Dbpedia spotlight: shedding light on the web of documents.
\newblock In \emph{Proceedings of the 7th international conference on semantic
  systems}, pages 1--8. ACM.

\bibitem[{Mikolov et~al.(2013)Mikolov, Sutskever, Chen, Corrado, and
  Dean}]{mikolov2013distributed}
Tomas Mikolov, Ilya Sutskever, Kai Chen, Greg~S Corrado, and Jeff Dean. 2013.
\newblock Distributed representations of words and phrases and their
  compositionality.
\newblock In \emph{Advances in neural information processing systems}, pages
  3111--3119.

\bibitem[{Moro et~al.(2014)Moro, Raganato, and Navigli}]{moro2014entity}
Andrea Moro, Alessandro Raganato, and Roberto Navigli. 2014.
\newblock Entity linking meets word sense disambiguation: a unified approach.
\newblock \emph{Transactions of the Association for Computational Linguistics},
  2:231--244.

\bibitem[{Nguyen et~al.(2016)Nguyen, Theobald, and Weikum}]{nguyen2016j}
Dat~Ba Nguyen, Martin Theobald, and Gerhard Weikum. 2016.
\newblock J-nerd: joint named entity recognition and disambiguation with rich
  linguistic features.
\newblock \emph{Transactions of the Association for Computational Linguistics},
  4:215--229.

\bibitem[{Nuzzolese et~al.(2015)Nuzzolese, Gentile, Presutti, Gangemi,
  Garigliotti, and Navigli}]{nuzzolese2015open}
Andrea~Giovanni Nuzzolese, Anna~Lisa Gentile, Valentina Presutti, Aldo Gangemi,
  Dar{\'\i}o Garigliotti, and Roberto Navigli. 2015.
\newblock Open knowledge extraction challenge.
\newblock In \emph{Semantic Web Evaluation Challenge}, pages 3--15. Springer.

\bibitem[{Piccinno and Ferragina(2014)}]{piccinno2014tagme}
Francesco Piccinno and Paolo Ferragina. 2014.
\newblock From tagme to wat: a new entity annotator.
\newblock In \emph{Proceedings of the first international workshop on Entity
  recognition \& disambiguation}, pages 55--62. ACM.

\bibitem[{Radhakrishnan et~al.(2018)Radhakrishnan, Talukdar, and
  Varma}]{radhakrishnan2018elden}
Priya Radhakrishnan, Partha Talukdar, and Vasudeva Varma. 2018.
\newblock Elden: Improved entity linking using densified knowledge graphs.
\newblock In \emph{Proceedings of the 2018 Conference of the North American
  Chapter of the Association for Computational Linguistics: Human Language
  Technologies, Volume 1 (Long Papers)}, volume~1, pages 1844--1853.

\bibitem[{Rizzo et~al.()Rizzo, van Erp, and Troncy}]{rizzo2014benchmarking}
Giuseppe Rizzo, Marieke van Erp, and Rapha{\"e}l Troncy.
\newblock Benchmarking the extraction and disambiguation of named entities on
  the semantic web.

\bibitem[{R{\"o}der et~al.(2017)R{\"o}der, Usbeck, and
  Ngonga~Ngomo}]{roder2017gerbil}
Michael R{\"o}der, Ricardo Usbeck, and Axel-Cyrille Ngonga~Ngomo. 2017.
\newblock Gerbil--benchmarking named entity recognition and linking
  consistently.
\newblock \emph{Semantic Web}, (Preprint):1--21.

\bibitem[{Sil and Yates(2013)}]{sil2013re}
Avirup Sil and Alexander Yates. 2013.
\newblock Re-ranking for joint named-entity recognition and linking.
\newblock In \emph{Proceedings of the 22nd ACM international conference on
  Conference on information \& knowledge management}, pages 2369--2374. ACM.

\bibitem[{Speck and Ngomo(2014)}]{speck2014ensemble}
Ren{\'e} Speck and Axel-Cyrille~Ngonga Ngomo. 2014.
\newblock Ensemble learning for named entity recognition.
\newblock In \emph{International semantic web conference}, pages 519--534.
  Springer.

\bibitem[{Spitkovsky and Chang()}]{spitkovsky2012cross}
Valentin~I Spitkovsky and Angel~X Chang.
\newblock A cross-lingual dictionary for english wikipedia concepts.

\bibitem[{Steinmetz and Sack(2013)}]{steinmetz2013semantic}
Nadine Steinmetz and Harald Sack. 2013.
\newblock Semantic multimedia information retrieval based on contextual
  descriptions.
\newblock In \emph{Extended Semantic Web Conference}, pages 382--396. Springer.

\bibitem[{Sun et~al.(2015)Sun, Lin, Tang, Yang, Ji, and Wang}]{sun2015modeling}
Yaming Sun, Lei Lin, Duyu Tang, Nan Yang, Zhenzhou Ji, and Xiaolong Wang. 2015.
\newblock Modeling mention, context and entity with neural networks for entity
  disambiguation.
\newblock In \emph{Twenty-Fourth International Joint Conference on Artificial
  Intelligence}.

\bibitem[{Yamada et~al.(2016)Yamada, Shindo, Takeda, and
  Takefuji}]{yamada2016joint}
Ikuya Yamada, Hiroyuki Shindo, Hideaki Takeda, and Yoshiyasu Takefuji. 2016.
\newblock Joint learning of the embedding of words and entities for named
  entity disambiguation.
\newblock In \emph{Proceedings of The 20th SIGNLL Conference on Computational
  Natural Language Learning}, pages 250--259.

\bibitem[{Yang et~al.(2018)Yang, Irsoy, and Rahman}]{yang2018collective}
Yi~Yang, Ozan Irsoy, and Kazi~Shefaet Rahman. 2018.
\newblock Collective entity disambiguation with structured gradient tree
  boosting.
\newblock In \emph{Proceedings of the 2018 Conference of the North American
  Chapter of the Association for Computational Linguistics: Human Language
  Technologies, Volume 1 (Long Papers)}, volume~1, pages 777--786.

\end{thebibliography}
